%% file: main.tex
\pdfoutput=1
\documentclass[11pt]{article}

\usepackage[final]{acl}

\usepackage{times}
\usepackage{latexsym}
\usepackage[T1]{fontenc}
\usepackage[utf8]{inputenc}
\usepackage{microtype}
\usepackage{inconsolata}

\usepackage{graphicx}
\graphicspath{{figs/}}            % all figures live in figs/
\usepackage{subcaption}
\usepackage{float}               % [H] placement: keep Section 3 figures in-flow
\usepackage{placeins}            % \FloatBarrier to pin section-end figures
\usepackage{booktabs}
\usepackage{multirow}
\usepackage{array}
\usepackage{amsmath}
\usepackage{amssymb}
\usepackage{xcolor}
\usepackage{url}
\usepackage{algorithm}
\usepackage{algpseudocode}

\usepackage[capitalize,nameinlink]{cleveref}
\crefname{algorithm}{Algorithm}{Algorithms}
\Crefname{algorithm}{Algorithm}{Algorithms}

\newcommand{\TQ}{TurboQuant}
\newcommand{\TQfour}{TQ$_{4/4}$}
\newcommand{\UQ}{UltraQuant}
\newcommand{\FPeight}{FP8}
\newcommand{\BFsixteen}{BF16}

\title{UltraQuant: 4-bit KV Caching for\\
Context-Heavy Agents}

\author{
  \mdseries
  Inesh Chakrabarti\thanks{Equal contribution.}$^{,1,2}$ \quad David Limpus$^{*,1,3}$ \quad Aditi Ghai Rana$^{*,1}$ \\
  Bowen Bao$^1$ \quad Spandan Tiwari$^1$ \quad Thiago Crepaldi$^1$ \quad Ashish Sirasao$^1$ \\[0.5em]
  $^1$Advanced Micro Devices \quad
  $^2$University of California, Los Angeles \quad
  $^3$Purdue University \\[0.3em]
  \texttt{inesh33@g.ucla.edu} \quad \texttt{dlimpus@purdue.edu} \\
  \texttt{\{aditi.rana, bowen.bao, spandan.tiwari, thiago.crepaldi, ashish.sirasao\}@amd.com}
}

\date{August 28, 2026}

\begin{document}
\maketitle

% --- Abstract ----------------------------------------------------------------
\input{sections/00_abstract}

% --- Body --------------------------------------------------------------------
\input{sections/01_introduction}
\input{sections/02_related_work}
\input{sections/03_agentic_workloads}
\input{sections/03_ultra_turboquant}

\input{sections/04c_ultra_quant}
\input{sections/06_accuracy_results}
\input{sections/06_performance_results}
\input{sections/07_conclusion}
\input{sections/06c_limitations}

% --- References --------------------------------------------------------------
% Bibliography style is set inside acl.sty; do not call \bibliographystyle.
\bibliography{references/references,references/anthology}

% --- Appendix ----------------------------------------------------------------
\appendix
\input{sections/05_kernel_implementation}
% \input{appendix/C_kernel_details}  % Include after update to latest flydsl details
% appendix/B_additional_results merged into G_accuracy_supplement
\input{appendix/D_per_block_ablation}        % B: Ablations
\input{appendix/G_accuracy_supplement}        % C: Additional Accuracy Results
\input{appendix/H_synthetic_multiturn}        % D: Synthetic multi-turn ablation
\input{appendix/F_claude_trace_adaptive}      % E: Claude Code trace replay
\input{appendix/E_turboquant_algorithm}       % F: Ultra-TQ LUT encoding

\end{document}

%% file: sections/00_abstract.tex
\begin{abstract}
Context-heavy agents place substantial pressure on the key--value (KV)
cache: long prefixes are reused across many short turns, while
concurrency determines whether the serving system can keep GPUs
utilized. We study 4-bit KV-cache compression for this setting, using
\TQ{}-style rotation and codebook quantization~\citep{zandieh2026turboquant} as a quality anchor and
vLLM \FPeight{} KV caching as the deployment anchor. We report three
contributions. First, we frame 4-bit KV caching around multi-round
agent workloads where task quality, cache residency, and serving
throughput must be measured jointly. Second, we describe the practical
design choices needed to make the 4-bit path robust, including
asymmetric K/V treatment, Walsh--Hadamard rotation~\citep{walsh1923closed,hadamard1893resolution}, QJL removal~\citep{zandieh2024qjl}, and
block-scale variants. Third, we present serving optimizations on AMD
GPUs, including optimized decode-attention kernels and
\emph{UltraQuant}, an FP4 approximation path that uses FP8 queries, FP4
KV tensors, UE8M0 group scales, and native scaled-MFMA support on
CDNA4. On an adaptive-SLO replay of production Claude Code traces,
\UQ{} sustains $2.71\times$ (MiniMax-M2.5) and $4.38\times$
(Qwen3-235B) the qualified-request throughput of the \BFsixteen{}
baseline, matching or exceeding hardware \FPeight{} KV while using half
the KV bytes. \UQ{} delivers its largest gains in long-context,
high-concurrency, memory-constrained serving regimes.
\end{abstract}

%% file: sections/01_introduction.tex
\section{Introduction}
\label{sec:introduction}
\label{sec:background}

Large language models have evolved from chatbots with short context
windows into agents that browse the web, inspect repositories, invoke
tools, and complete substantial software-engineering tasks~\citep{yao2022react,schick2023toolformer,zhou2023webarena,yang2024sweagent}. These
workflows require long-running memory: the model must retain system
instructions, tool definitions, retrieved documents, codebase context,
and the evolving plan across many turns. As model releases push
context windows toward one million tokens and beyond~\citep{geminiteam2024gemini15,ding2024longrope}, the KV cache
grows linearly with context length and becomes a first-order consumer
of high-bandwidth memory (HBM).

This has driven two complementary lines of work: One line
changes the model architecture or attention mechanism to reduce KV
state directly, including multi-head latent attention~\citep{deepseekv2_2024,deepseekv3_2024} and linear or
state-space alternatives~\citep{gu2023mamba,dao2024mamba2,yang2024deltarule,yang2024gateddelta}. Another line treats the memory hierarchy as
a systems problem, using cache reuse, offloading, and paged-memory
management to keep scarce HBM occupied by the most useful state~\citep{kwon2023vllm,zheng2023sglang,zhang2024pyramidkv}. KV
quantization adopts the first approach: we preserve the standard interface while reducing memory overhead.

Compression leads to some challenges; a deployable 4-bit method must
offer more than just compression, but rather must retain answer quality on long-context
workloads, integrate with paged serving, and dispatch efficiently through GPU matrix cores.
\FPeight{} KV caching is  a strong deployment baseline in vLLM, giving roughly $2\times$ compression 
at near-lossless quality with native hardware support. \TQ{}~\citep{zandieh2026turboquant}, a
data-oblivious 4-bit scalar-plus-rotation method, provides a strong
4-bit quality point but relies on codebook lookup and software dequant.

We study two endpoints on AMD Instinct GPUs. \emph{Ultra-TQ} keeps the
\TQ{} representation and closes the kernel gap through layout, lookup,
and MFMA scheduling optimizations, along with accuracy improvements
through calibrated centroids for each model. \emph{UltraQuant} replaces
the codebook with an FP4 micro-tensor approximation so the
dequantization is folded into native CDNA4 scaled-MFMA instructions.

Our goal is a deployment-oriented account of low-bit KV caching for
context-heavy agents: which quantization choices preserve quality, how
those choices interact with decode-attention kernels, and when the
serving system benefits from the additional resident context. \UQ{}
delivers its largest gains in long-context, high-concurrency,
memory-constrained serving regimes.

%% file: sections/02_related_work.tex
\section{Related Work}
\label{sec:related-work}

\subsection{KV-Cache Quantization}
\label{sec:rw-kv}

The KV-compression landscape has converged on a few recurring design
ideas. Keys and values are treated asymmetrically because key errors
perturb the softmax distribution differently from value errors, a
pattern established by KIVI~\citep{liu2024kivi} and
KVQuant~\citep{hooper2024kvquant}. 
QJL~\citep{zandieh2024qjl} and
\TQ{}~\citep{zandieh2026turboquant} use rotation and codebook methods
to improve rate--distortion at small bit widths, and \TQ{} in particular
has seen community reimplementation~\citep{thetom2026turboquantplus,chakrabarti2026productionizing}. However,
the use of the codeboook introduces lookup and irregular-access costs \citep{chakrabarti2026productionizing}.
These methods are useful algorithmic anchors for 4-bit quality, but their most accurate
representations are not automatically efficient serving formats.

\subsection{Rotation and Codebook Quantization}
\label{sec:rw-rotation}

\TQ{}-style rotation~\citep{kurtic2026vllm} spreads outlier energy across channels so scalar
quantizers see a more Gaussian-like distribution. Calibrated
centroids fit via per-model Lloyd--Max~\citep{lloyd1982least,max1960quantizing} or its per-block variant (LMPB) then provide a
strong 4-bit codebook for the rotated distribution. In this paper,
those codebooks are background and motivation rather than the deployed
format: UltraQuant uses them to explain why FP4 is plausible
after rotation, then replaces arbitrary centroids with the
hardware-native FP4 grid.

\subsection{Serving Systems and Hardware-Native Low Precision}
\label{sec:rw-serving}

vLLM-style paged serving~\citep{kwon2023vllm} makes KV-cache residency
a systems concern, not only a compression metric. Hardware-native
formats such as \FPeight{} already benefit from direct matrix-core
support. UltraQuant extends this serving-first lens to 4-bit KV
caching by targeting the CDNA4 scaled-F8F6F4 MFMA path, where FP4
operands and UE8M0 scales are consumed by the matrix core itself.

%% file: sections/03_agentic_workloads.tex
\section{Agentic Serving on Production Traces}
\label{sec:agentic-workloads}

For context-heavy agents, each session keeps a long prefix resident in
HBM across many turns, so the number of agents a GPU can serve under a
latency service-level objective (SLO) depends on how much KV state fits
in memory. A smaller KV budget fills first, evicts reusable prefix
blocks, and forces expensive re-prefill that spikes time-to-first-token
(TTFT). A KV method must therefore keep long prefixes resident at high
concurrency and decode at near hardware-native speed, since
dequantization overhead can offset the capacity it buys.

We emulate production Claude Code workloads using anonymized,
production-derived sessions from the open-source
\texttt{kv-cache-tester} corpus~\citep{fox2024kvcachetester}. Virtual
users replay recorded sessions against our vLLM endpoint, reproducing
multi-turn order, prompt/output-token targets, inter-turn think times,
and prefix-sharing structure.

We measure serving under an adaptive-SLO protocol on MiniMax M2.5 and
Qwen3-235B. An admission controller ramps from 6 to 48 concurrent
sessions, gated by a rolling p95 TTFT against a 120-second target;
admission pauses when TTFT headroom drops below the gate. This mimics
how a production scheduler trades concurrency against latency. Our primary metric
is TTFT-qualified goodput: the rate of completed requests whose first
token arrives within the TTFT budget. We report cumulative SLO-qualified
requests in the main text; \cref{app:claude-trace-adaptive} gives
goodput curves and effective-TTFT tail distributions.

The key mechanism is capacity-driven prefix retention. \FPeight{} KV
ramps faster at low pressure but fills earlier, evicting reusable
prefix blocks and forcing re-prefill. UltraQuant tracks \FPeight{} KV
decode speed under low pressure while offering 3.2--3.4$\times$ BF16
logical KV capacity (vs.\ 2$\times$ for \FPeight{}), reaching
2.34$\times$/3.25$\times$ BF16's prefix-hit ratio on MiniMax/Qwen
(vs.\ 1.07$\times$/0.85$\times$ for \FPeight{}). The avoided prefill
work outweighs UltraQuant's slight decode overhead under pressure.

\begin{figure*}[t]
  \centering
  \begin{subfigure}{0.49\textwidth}
    \centering
    \includegraphics[width=\linewidth]{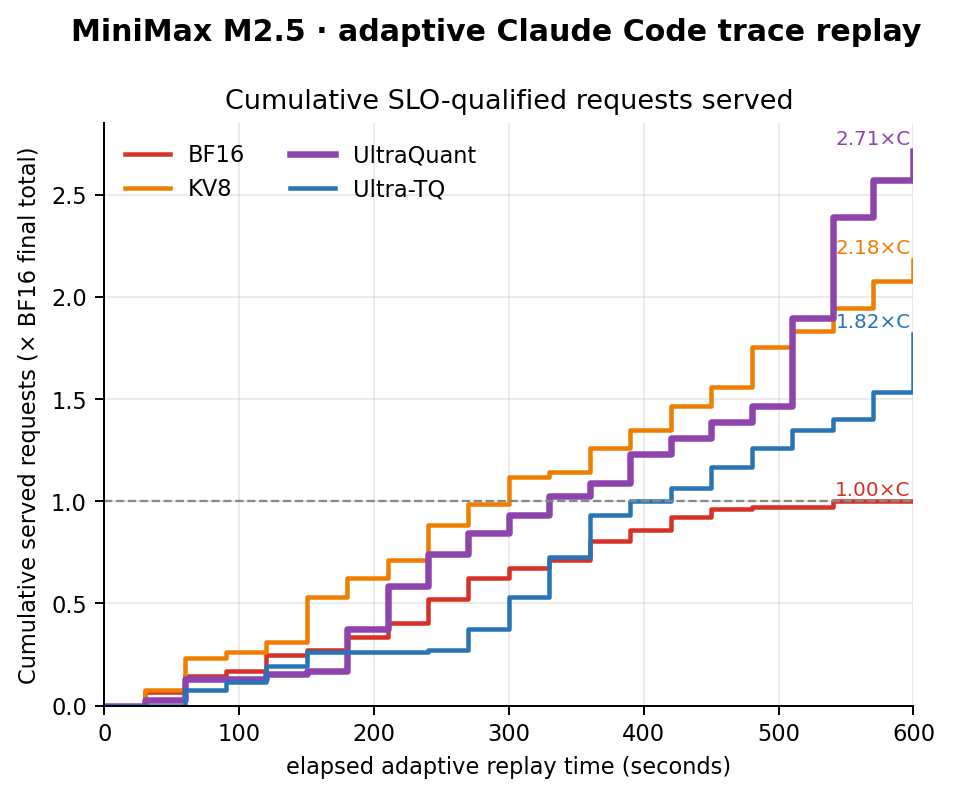}
    \caption{MiniMax M2.5.}
    \label{fig:trace-adaptive-minimax}
  \end{subfigure}\hfill
  \begin{subfigure}{0.49\textwidth}
    \centering
    \includegraphics[width=\linewidth]{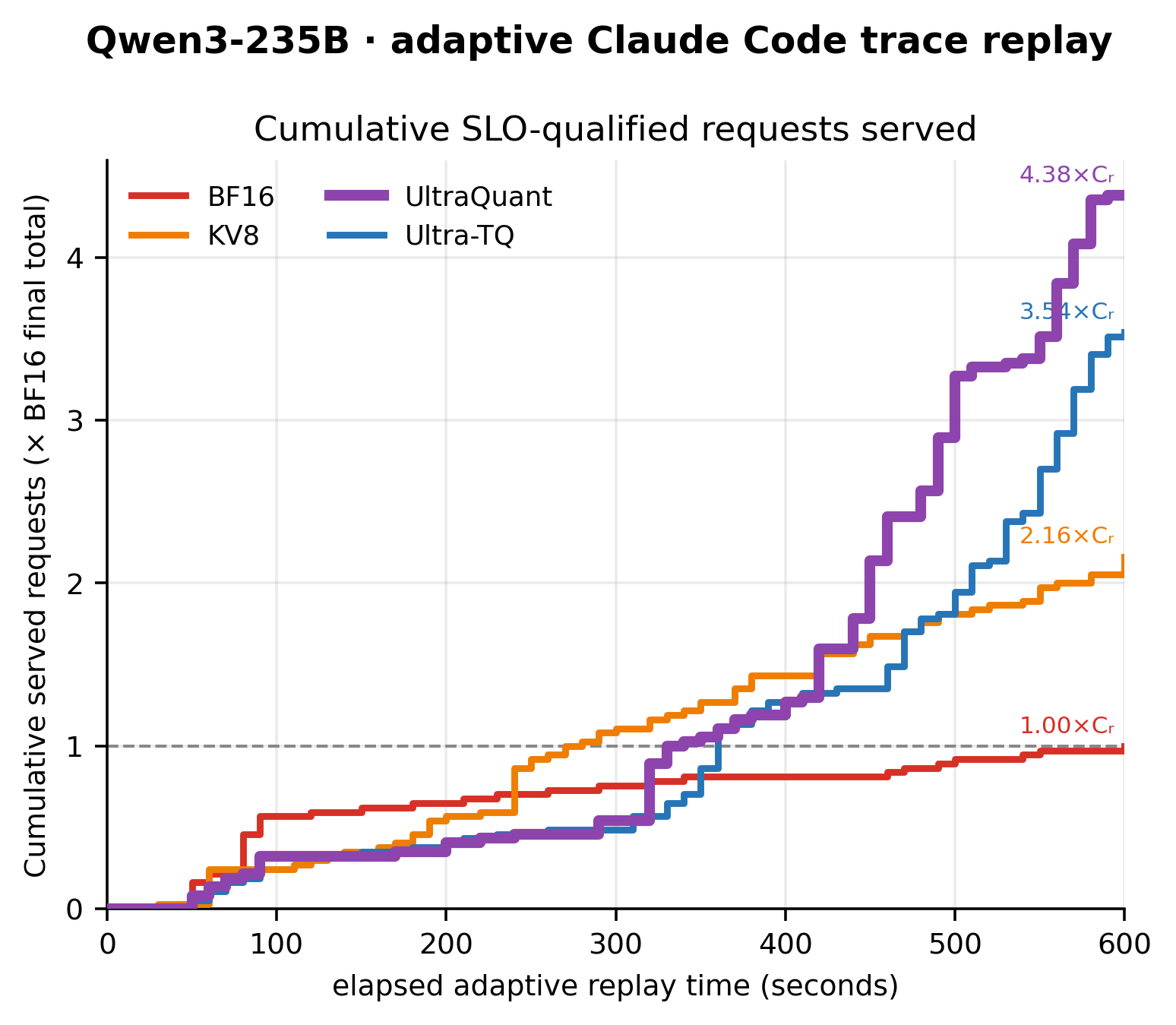}
    \caption{Qwen3-235B.}
    \label{fig:trace-adaptive-qwen}
  \end{subfigure}
  \caption{Cumulative SLO-qualified requests served on production-derived
    Claude Code traces (TP$=2$, AMD MI355X), normalized relative to
    \BFsixteen{}.}
  \label{fig:trace-adaptive}
\end{figure*}

\paragraph{MiniMax M2.5.}
\FPeight{} KV reaches the concurrency cap earlier, but UltraQuant's
SLO-qualified curve accelerates under pressure and finishes at
2.71$\times$ BF16's cumulative qualified-request total, compared with
2.18$\times$ for \FPeight{} KV and 1.82$\times$ for Ultra-TQ
(\cref{fig:trace-adaptive-minimax}).

\paragraph{Qwen3-235B.}
UltraQuant preserves more TTFT headroom as cache pressure rises,
finishing at 4.38$\times$ BF16's qualified-request total vs.\
2.16$\times$ for \FPeight{} KV and 3.54$\times$ for Ultra-TQ
(\cref{fig:trace-adaptive-qwen}).

\Cref{fig:trace-adaptive-minimax,fig:trace-adaptive-qwen} normalize
cumulative SLO-qualified turns by \BFsixteen{}'s qualified-request
total. Across both models, \FPeight{} KV has lower compute overhead at
low pressure, while UltraQuant retains more KV state under high
pressure. Once prefix replacement and re-prefill become expensive,
UltraQuant's capacity advantage outweighs its dequant cost.
\Cref{tab:agentic-trace-goodput} reports the terminal TTFT-qualified
goodput at the 120-second threshold on the same replay: UltraQuant
reaches $2.71\times$/$4.38\times$ \BFsixteen{}'s qualified-request rate
on MiniMax and Qwen. Aggregate throughput confirms this: UltraQuant
reaches 1.25$\times$/1.29$\times$ BF16's input/output tok/s on MiniMax
and 1.59$\times$/1.53$\times$ on Qwen. Full TTFT-goodput curves and
effective-TTFT tail panels for the same Claude Code replay are in
\cref{app:claude-trace-adaptive}. The practical takeaway is that on
identical hardware and under a fixed 120-second SLO, 4-bit KV serves
$2.7$--$4.4\times$ more agent requests than BF16 while matching or
exceeding hardware \FPeight{} KV at half the cache footprint.

\begin{table}[t]
  \centering
  \scriptsize
  \setlength{\tabcolsep}{3pt}
  \resizebox{\columnwidth}{!}{%
  \begin{tabular}{llrr}
    \toprule
    \textbf{Model} & \textbf{Scheme} &
    \textbf{Qual.\ @120s} &
    \textbf{Goodput (req/s/GPU)} \\
    \midrule
    \multirow{4}{*}{MiniMax M2.5}
      & \BFsixteen{}     & 77  & 0.064 \\
      & \FPeight{} KV    & 168 & 0.140 \\
      & Ultra-TQ         & 140 & 0.117 \\
      & \UQ{}            & 209 & 0.174 \\
    \midrule
    \multirow{4}{*}{Qwen3-235B}
      & \BFsixteen{}     & 37  & 0.031 \\
      & \FPeight{} KV    & 80  & 0.067 \\
      & Ultra-TQ         & 131 & 0.109 \\
      & \UQ{}            & 162 & 0.135 \\
    \bottomrule
  \end{tabular}
  }
  \caption{TTFT-qualified goodput on the production-derived Claude Code
    adaptive trace replay (600~s assessment window, TP$=2$, AMD MI355X).
    A completion qualifies if its effective TTFT is $\le 120$~s; goodput
    is the qualifying count divided by the assessment window and GPU
    count.}
  \label{tab:agentic-trace-goodput}
\end{table}

%% file: sections/03_ultra_turboquant.tex
\section{Ultra-TurboQuant}
\label{sec:ultra-turboquant}

Ultra-TurboQuant (Ultra-TQ) is our optimized implementation of
\TQfour{}, the 4-bit \TQ{} KV-cache scheme. It starts from the same observation that motivates \TQ{}: the
raw KV-cache distribution is a poor target for a small scalar codebook,
but a fixed Walsh--Hadamard rotation spreads outlier energy across
channels, pushing each coordinate toward the light-tailed,
near-Gaussian regime where a 4-bit scalar quantizer is a reasonable
approximation. \Cref{sec:kernels} presents efficient implementation details.

\subsection{Calibrated Centroids}
\label{sec:utq-calibrated-centroids}

We begin with the observation that calibrating TurboQuant centroids is
model agnostic. The production Lloyd--Max centroids assume post-rotation, post-normfold key
coordinates follow $\mathrm{Beta}((d-1)/2, (d-1)/2)$, the exact law of a
randomly rotated unit-vector coordinate. On real activations this is only approximate,
so the theoretical centroids are slightly
misplaced. Refitting the 16-entry Lloyd--Max table on captured key
activations corrects this. Calibration is cheap --- a single forward
pass over ${\sim}20$ randomly sampled vectors from a sphere per rotated layer yields 16 FP32 levels ---
and deployment is a drop-in table swap with zero inference overhead. In practice,
we only refit 10\% of layers that show higher per-element quantizaiton MSE. 

Calibration is the strongest pure-codebook lever we measured: it lowers
per-element K quantization MSE by $10.3\%$
($1.32\times10^{-4} \rightarrow 1.18\times10^{-4}$) and improves GPQA by
$+1.20$\,pp. The empirical centroids are compressed relative to the
theoretical ones (largest $|c|$ at $d = 64$ falls from $0.342$ to
$0.288$), reflecting the slight deviation of real activations from the
ideal. Rotation remains
essential: on the un-rotated distribution,
even model-tuned centroids leave ${\sim}2.3\times$ the rotated MSE,
suggesting that centroid placement cannot
replicate rotation's whitening effect. The gain from calibrating
centroids is real but modest; we ablate the source of the benefits in
\cref{app:per-block-ablation}.

%% file: sections/04c_ultra_quant.tex
\section{UltraQuant}
\label{sec:ultra-quant}

A calibrated codebook must be decoded at every attention step: each
tile loads packed indices, gathers centroids through a lookup table,
and reconstructs keys/values in registers before the QK and PV
products~\citep{guo2024flute}. This gather is irregular and grows with
context length. UltraQuant instead stores the cache as MXFP4: FP4 E2M1 codes
with UE8M0 scales that the matrix core reads natively, making
dequantization a single exponent shift folded into the instruction.
The per-group scales is derived from one global constant, calibrated
offline and model-agnostic, that minimizes quantization error on the
rotated distribution.

\subsection{Cache Layout}
\label{sec:uq-cache-layout}

UltraQuant stores keys/values in MXFP4. The head dimension is partitioned into contiguous groups of 32
channels, each encoded as 32 FP4 E2M1 codes (packed two per byte) plus
one UE8M0 scale: $32 {\times} 4\text{ bits} + 8\text{ bits} = 17\text{ B}$
per group, i.e.\ $4.25$ bits per element --- within $6\%$ of an ideal
4-bit representation, in a format the matrix core consumes natively.

\subsection{Dequantization}
\label{sec:uq-dequant}

The dequantization rule is:
\begin{equation}
  \texttt{value} = \texttt{code} \times 2^{\texttt{scale}}, \qquad
  \texttt{scale} = \texttt{byte} - 127 .
\end{equation}
The scale is a power-of-two exponent that shifts the FP4 codepoint and
folds directly into the scaled-MFMA accumulator. The AMD
\texttt{MFMA\_SCALE\_F32\_*\_F8F6F4} instruction takes FP4 codes and
the UE8M0 byte as native operands~\citep{amd2024cdna4isa}, so keys and
values are never materialized in BF16. Queries are Hadamard-rotated and
rounded to FP8 E4M3, so attention runs on the FP8$\times$FP4
scaled-MFMA path. The constant $c$ (\cref{sec:uq-constopt}) enters
through the encoder rule $E = \mathrm{round}(\log_2(c \cdot m))$ rather
than as a runtime multiply.

\subsection{Constant-Optimized Scaling}
\label{sec:uq-constopt}

\begin{figure*}[t]
  \centering
  \begin{subfigure}{0.49\textwidth}
    \centering
    \includegraphics[width=\linewidth]{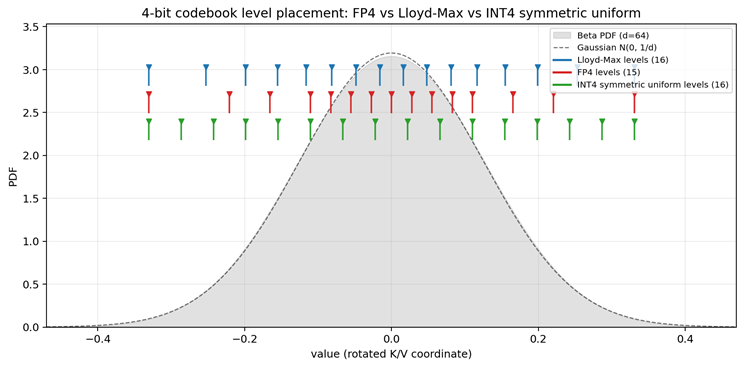}
    \caption{4-bit codebook level placement at $d{=}64$.}
    \label{fig:codebook-placement}
  \end{subfigure}\hfill
  \begin{subfigure}{0.49\textwidth}
    \centering
    \includegraphics[width=\linewidth]{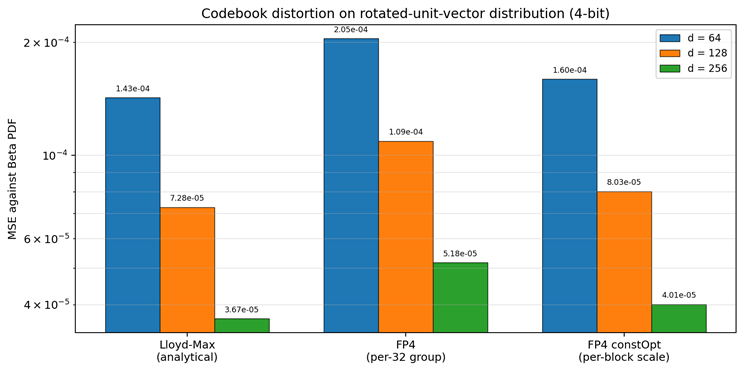}
    \caption{Codebook distortion (MSE).}
    \label{fig:codebook-distortion}
  \end{subfigure}
  \caption{Codebook analysis on the rotated-unit-vector distribution.
    (a)~Lloyd--Max centroids provide the algorithmic reference, FP4 E2M1
    is the hardware-native approximation, and symmetric INT4 is a
    uniform baseline. (b)~FP4 has higher MSE than analytical Lloyd--Max,
    but constOpt per-block scaling closes much of the gap while
    preserving a hardware-native FP4 representation.}
  \label{fig:codebook}
\end{figure*}

Lloyd--Max centroids for the rotated distribution are symmetric,
concentrated near zero, and increasingly wide in the tails: the
pattern FP4 must approximate. \cref{fig:codebook-placement} shows
Lloyd--Max placing levels more densely in the high-probability body
than uniform INT4, while FP4 E2M1 has fixed hardware-defined spacing.
Plain per-group FP4 is consistently worse than Lloyd--Max in
MSE~\citep{dalberto2026statistical}, but the constant-optimized
per-block scale below recovers most of that gap while keeping the FP4
grid, as quantified in \cref{fig:codebook-distortion} and ablated in
\cref{app:global-constant-ablation}.

Each group of 32 channels is mapped onto the FP4 grid by a single scale
$s = c \cdot m$, where $m = \max_i |x_i|$ is the group's absolute
maximum and $c$ is one global constant shared across every group, head,
and model. We drop the per-token $\ell_2$ normalization (K-norm) used by
the codebook path: each block's scale derives from its own absmax,
making the blocks self-scaling. A per-token factor cannot be folded into
the attention score (it does not pass through the softmax), so it would
add a stored scalar and runtime multiply with no quality gain.

The per-group absmax $m$ absorbs all per-token, per-head, and per-model
magnitude variation: dividing by $m$ maps each group into $[-1,1]$.
After Hadamard rotation the remaining \emph{shape} is approximately
$\mathrm{Beta}((d{-}1)/2, (d{-}1)/2)$, depending only on head
dimension~$d$. This model-independence is why a single constant works:
the one remaining degree of freedom is where this fixed shape sits
against the E2M1 grid, i.e.\ the ratio $c = s/m$.
In effect, the per-group absmax together with the constant
$c$ takes over the role of TurboQuant's per-token K-norm.

We choose $c$ to minimize the reconstruction error of the E2M1 grid.
With $q(\cdot)$ the nearest-codepoint rounding onto
$F = \{0, \pm 0.5, \pm 1, \pm 1.5, \pm 2, \pm 3, \pm 4, \pm 6\}$:
\begin{equation}
  c^{*} = \arg\min_{c}\;
  \mathbb{E}\!\left[\bigl(z - c\,m\,q(z / (c\,m))\bigr)^{2}\right] .
\end{equation}
This one-dimensional minimization is solved once, offline, on captured
rotated key activations. The minimizer is $c = 0.156$, deployed without
per-model recalibration across MiniMax and Qwen
(\cref{sec:accuracy-results}); it remains MSE-optimal in the
single-model ablation in \cref{app:global-constant-ablation}.
\Cref{fig:codebook-distortion} shows the codebook-level MSE gap.

This one-dimensional minimization is solved once, offline,
on captured rotated key activations over a small calibration set.
The minimizer on the captured keys is
$c = 0.156$, the value we deploy. Note that the distribution post rotation is similar
 enough that this single fit transfers across models at a given head:
 we deploy the same $c = 0.156$, without any per-model recalibration,
 across the MiniMax and Qwen model families reported in
 \cref{sec:accuracy-results}, and it remains the MSE-optimal setting
 in the single-model ablation in \cref{app:global-constant-ablation}.

To ablate, we consider alternative constants in
\Cref{app:global-constant-ablation}; \Cref{fig:codebook-distortion}
shows the corresponding codebook-level MSE gap.

%% file: sections/06_accuracy_results.tex
\section{Accuracy Results}
\label{sec:accuracy-results}

All KV-compressed evaluations use boundary-layer protection: the first and
last two attention layers retain \BFsixteen{} KV caches ($n{=}2$). Our
primary accuracy result is end-to-end agent task resolution on SWE-bench
Lite (\cref{sec:accuracy-swebench}): it is the only benchmark here that
grades a complete multi-turn agent loop rather than a single response, and
it is the setting UltraQuant targets, since a long shared repository prefix
with many follow-up turns is exactly where KV capacity binds.
\Cref{sec:accuracy-retrieval,sec:accuracy-reasoning} summarize retrieval and
reasoning accuracy; their per-model tables and per-task breakdowns are in
\cref{app:accuracy-supplement}.

\subsection{SWE-bench Lite Agent Tasks}
\label{sec:accuracy-swebench}

SWE-bench Lite evaluates end-to-end software-engineering task resolution:
for each GitHub issue, an agent inspects the repository, edits over multiple
turns, and submits a patch graded by project tests. We evaluate every KV
format on a fixed 100-task cohort for MiniMax~M2.5 and Qwen3-235B with
prefix caching enabled (\cref{tab:accuracy-swebench}).

\begin{table}[t]
  \centering
  \scriptsize
  \resizebox{\columnwidth}{!}{%
  \begin{tabular}{lrrrrr}
    \toprule
    \textbf{Model} & \textbf{\BFsixteen{}} & \textbf{\FPeight{} KV} &
    \textbf{Ultra-TQ} & \textbf{UltraQuant} &
    \textbf{$\Delta$~BF16} \\
    \midrule
    MiniMax M2.5 & \textbf{62} & \textbf{62} & 61 & 59 & $-3$ \\
    Qwen3-235B   & \textbf{28} & 21 & 27 & 24 & $-4$ \\
    \bottomrule
  \end{tabular}
  }
  \caption{SWE-bench Lite resolved tasks (fixed 100-task denominator).}
  \label{tab:accuracy-swebench}
\end{table}

On MiniMax, UltraQuant resolves 59/100 tasks versus 62/100 for \BFsixteen{},
retaining 95\% of \BFsixteen{}'s resolved-task count, with \FPeight{} KV at
62/100 and Ultra-TQ at 61/100. On Qwen, UltraQuant resolves 24/100 versus
28/100 for \BFsixteen{}; it exceeds \FPeight{} KV's 21/100 and stays within
three points of Ultra-TQ's 27/100. Agent task resolution therefore lands
within a few points of \BFsixteen{} on both models while the KV cache is
stored at 4 bits, and on Qwen both 4-bit schemes beat the 8-bit baseline
despite its larger per-token budget.
\Cref{fig:accuracy-swebench} in \cref{app:accuracy-supplement} plots
resolved-task counts for the same cohort under all four KV schemes.

\subsection{Retrieval Tasks}
\label{sec:accuracy-retrieval}

Retrieval accuracy is largely preserved and the residual loss is
task-dependent rather than uniform. On Llama-3.1-8B-Instruct, UltraQuant
matches \BFsixteen{} on NIAH and on the 13-task LongBench-E average. Using
NVIDIA's official RULER suite (4K--131K, four tasks), Qwen3.5-35B-A3B reaches
exact parity at $99.38$, while MiniMax~M2.5 gives up ground at $74.03$ versus
$67.08$, with the loss concentrated in \texttt{single\_2} and
\texttt{multikey\_1} and a gain on \texttt{multiquery}
(\cref{app:accuracy-supplement}).

\subsection{Reasoning Tasks}
\label{sec:accuracy-reasoning}

We evaluate every KV format on GPQA Diamond, GSM8K, and AIME25. Deltas are
small in both directions: UltraQuant stays within $2$~points of \BFsixteen{}
on every model--benchmark pair except Qwen3-32B GPQA Diamond, where it gains
$5.58$ points. On MiniMax~M2.5 GPQA Diamond the intermediate schemes fall between
the two endpoints, with \FPeight{} KV at $83.33$ and Ultra-TQ matching
\BFsixteen{} at $83.84$ (\cref{fig:accuracy-reasoning-bfuq}).
MiniMax~M2.5 AIME25 is averaged over five seeds;
Llama-3.1-8B-Instruct AIME25 is omitted for insufficient coverage.

\FloatBarrier

%% file: sections/06_performance_results.tex
\section{Performance Results}
\label{sec:performance-results}

\begin{figure}[t]
  \centering
  \includegraphics[width=\columnwidth]{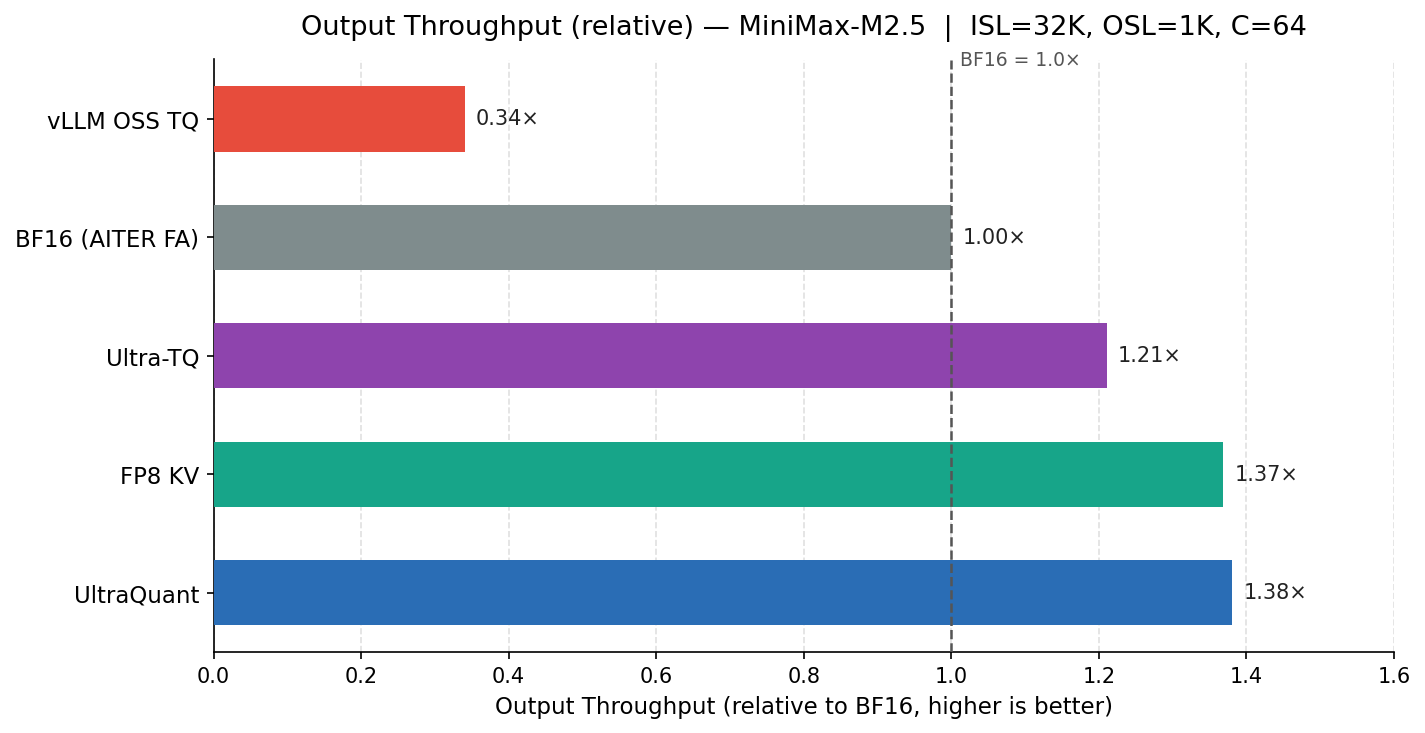}
  \caption{UltraQuant throughput (relative to \BFsixteen{}) vs.\
    \BFsixteen{}, \FPeight{} KV, and Ultra-TQ on MiniMax-M2.5
    (TP$=2$, $2\times$MI355X).}
  \label{fig:throughput-results}
\end{figure}

We measure UltraQuant's decode throughput and per-token latency
against four baselines: vLLM OSS \TQ{}~\citep{kurtic2026vllm}, \BFsixteen{} AITER~\citep{amd2025aiter}
FlashAttention, hardware \FPeight{} KV, and Ultra-TQ, on
MiniMax-M2.5, TP$=2$, $2\times$MI355X, 32K/1K at concurrency $64$. All
results are reported relative to the \BFsixteen{} AITER FlashAttention
baseline ($=1.00\times$). Folding dequantization into the MFMA gives
UltraQuant nearly the same instruction mix as the \FPeight{} kernel,
so it tracks the \FPeight{} throughput ceiling while halving the
KV-cache footprint. \Cref{fig:throughput-results} reports throughput
across the kernels: at $C{=}64$ UltraQuant reaches $1.38\times$ the
\BFsixteen{} output throughput, matching hardware \FPeight{} KV
($1.37\times$) to within ${\sim}1\%$, while serving twice the KV
footprint per unit of HBM.

\begin{figure*}[t]
  \centering
  \begin{subfigure}{0.49\textwidth}
    \centering
    \includegraphics[width=\linewidth]{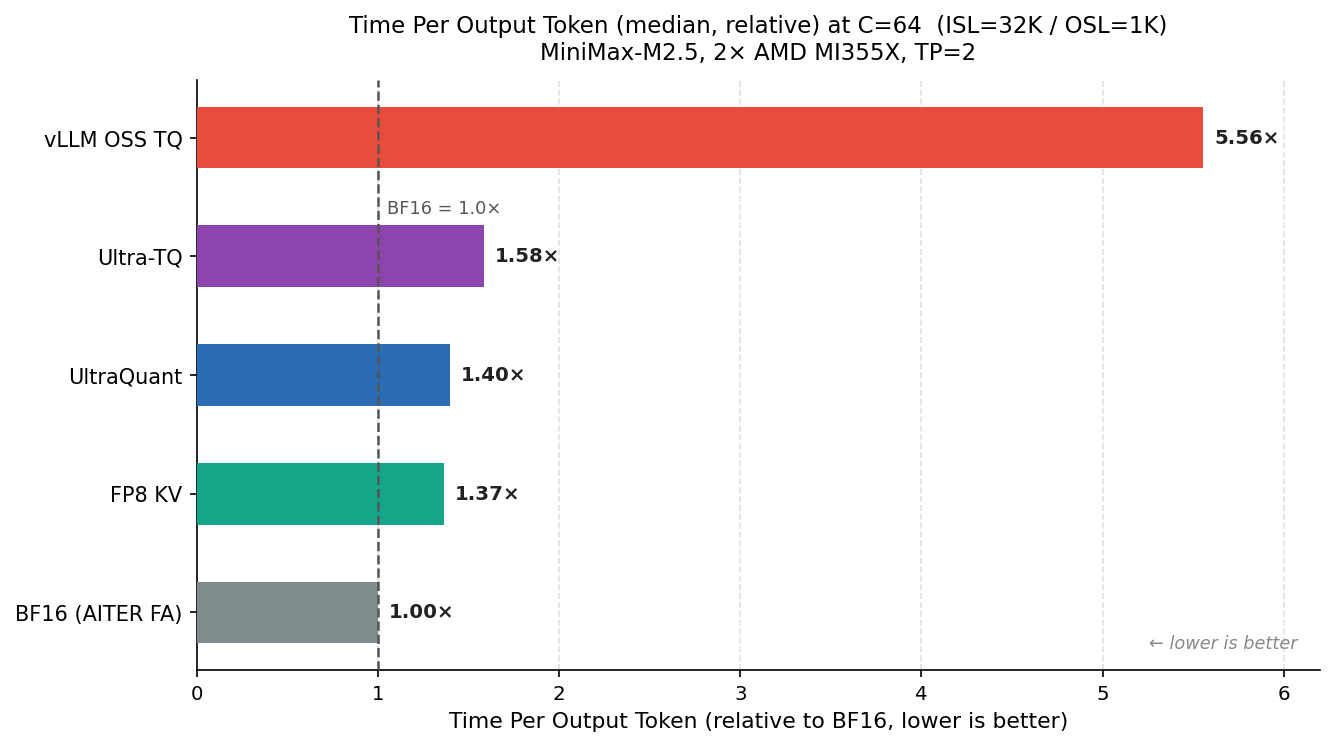}
    \caption{Median time per output token, $C{=}64$.}
    \label{fig:tpot-results}
  \end{subfigure}\hfill
  \begin{subfigure}{0.49\textwidth}
    \centering
    \includegraphics[width=\linewidth]{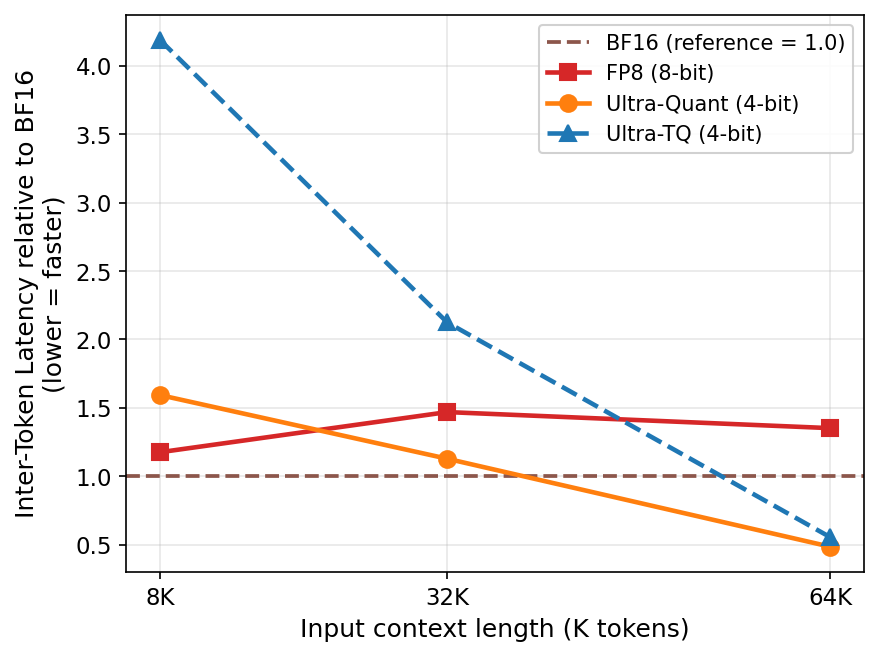}
    \caption{Inter-token latency vs.\ input context length.}
    \label{fig:itl-context}
  \end{subfigure}
  \caption{Per-token latency on MiniMax-M2.5 (32K/1K, TP$=2$,
    $2\times$MI355X), relative to \BFsixteen{} (lower is better) for
    \FPeight{}, \UQ{}, and Ultra-TQ.}
  \label{fig:latency-results}
\end{figure*}

The FP4 path largely removes software dequantization from the critical
path while retaining the memory-capacity benefit of a 4-bit KV cache.
UltraQuant delivers $1.38\times$ the \BFsixteen{} baseline throughput,
on par with the hardware \FPeight{} KV-cache path ($1.37\times$) while
using half the KV bytes per element. On per-token latency, \BFsixteen{}
has no dequant, so all 4-bit kernels sit above it; UltraQuant's median
TPOT is $1.40\times$ \BFsixteen{}, within ${\sim}2\%$ of the hardware
\FPeight{} KV-cache path ($1.37\times$) and well ahead of
Ultra-TQ ($1.58\times$) and vLLM OSS \TQ{} ($5.56\times$)
(\cref{fig:tpot-results}).

The advantage grows with context length.
\Cref{fig:itl-context} sweeps inter-token latency against input context
from 8K to 64K: \UQ{} improves monotonically relative to \BFsixteen{},
reaching ${\sim}0.5\times$ at 64K, while \FPeight{} stays near
$1.3$--$1.5\times$ and Ultra-TQ only catches up at the longest
contexts. The half-precision KV footprint is what lets \UQ{} sustain
longer contexts before HBM pressure forces re-prefill.

\subsection{Sensitivity to Concurrency, Input Length, and Output Length}
\label{sec:sensitivity-ablations}

The $C{=}64$ results above fix concurrency, input length, and output
length simultaneously. We additionally vary each axis independently
on MiniMax-M2.5, holding the others fixed. \UQ{} tracks \FPeight{} KV
closely on raw throughput and tail latency as concurrency and output
length grow. At long input context, \FPeight{} KV hits its
resident-cache memory wall around $40$K context and begins
evicting/recomputing sequences, while \UQ{} throughput and TTFT stay
flat through the same range: 4-bit KV buys capacity, not kernel speed,
and serves the long-context load that makes \FPeight{} KV fall behind.
The same input-length sweep on Qwen3-235B-A22B shows a consistent
trend.
Full per-axis figures,
discussion, and captions are in \cref{app:sensitivity-ablations}.

\subsection{Kernel-Level Roofline Analysis}
\label{sec:roofline}

Two mechanisms explain why \UQ{} tracks \FPeight{} on raw throughput
and latency while pulling ahead at long context and high load. First,
4-bit KV approximately doubles nominal cache capacity relative to
\FPeight{},
reducing eviction and re-prefill under pressure
(\cref{sec:sensitivity-ablations}). Second, long-context decode is
bandwidth-bound, so reading 4-bit KV moves about half as many bytes;
at short context or decode-heavy configurations, \FPeight{} can be
faster because the dequantization overhead is not yet amortized.

\begin{figure}[t]
  \centering
  \includegraphics[width=\columnwidth]{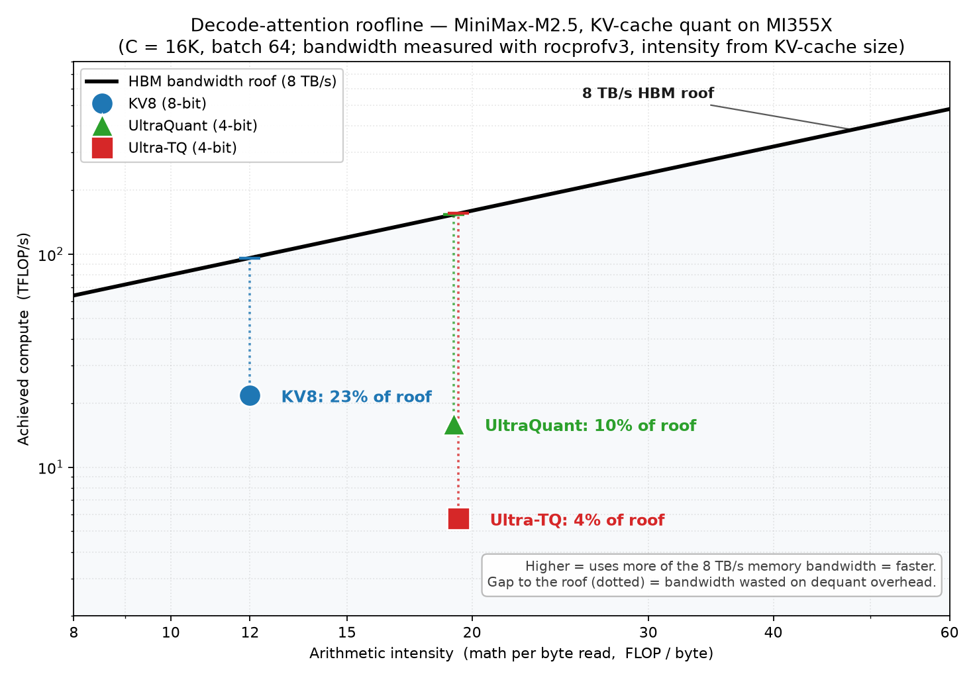}
  \caption{Decode-attention roofline on MiniMax-M2.5 (MI355X,
    $C{=}16$K, batch $64$): achieved compute vs.\ arithmetic intensity
    for \FPeight{} KV, \UQ{}, and Ultra-TQ against the $8$\,TB/s HBM
    bandwidth roof.}
  \label{fig:roofline}
\end{figure}

\Cref{fig:roofline} makes the kernel-level mechanism explicit with a
decode-attention roofline on MiniMax-M2.5 (MI355X, $C{=}16$K, batch
$64$). The horizontal axis is arithmetic intensity (math per byte
read); \FPeight{} KV sits furthest left because 8-bit KV moves the
most data per unit of math, while \UQ{} and Ultra-TQ shift right by
${\sim}1.6\times$ rather than the nominal $2\times$, since metadata
and padding leave a real footprint of about $5$ effective bits per
element. The vertical axis is achieved TFLOP/s against the
bandwidth-implied compute ceiling: \FPeight{} KV reaches $23\%$ of
that ceiling, \UQ{} reaches $10\%$, and Ultra-TQ reaches only $4\%$.

The winning condition is reaching 4-bit intensity while staying close
to bandwidth-limited rather than dequant-limited: \UQ{} reaches 4-bit
intensity and stays near bandwidth-bound, ${\sim}2.5\times$ higher than
Ultra-TQ on the roofline, whose software codebook decode is
dequant-bound. Kernel counters at $32$K context confirm \UQ{} retains
most of the memory saving at near-\FPeight{} kernel time; the remaining
${\sim}18\%$ gap is why its largest end-to-end wins come from capacity
and avoided re-prefill (\cref{sec:sensitivity-ablations}) rather than a
universal per-token speedup. \Cref{app:roofline-detail} gives the
per-scheme breakdown and counter table.

%% file: sections/07_conclusion.tex
\section{Conclusions}
\label{sec:conclusion}

We presented a 4-bit KV-cache study centered on context-heavy agents,
where long prefixes must remain reusable and concurrency determines GPU
utilization. \TQ{}-style rotation and codebook methods provide a strong
4-bit quality anchor; Ultra-TQ closes the gap to production serving through
careful layout, LUT, and MFMA scheduling; and UltraQuant replaces the codebook
with MXFP4 tensors consumed directly by CDNA4
matrix cores. The broader lesson is that low-bit KV caching should be
evaluated as an end-to-end serving mechanism, not only as an offline compression
method. For agent workloads, the
relevant outcome is the joint effect on task success, cache residency,
throughput, and interactivity. Our results show that 4-bit KV caching retains the
context-capacity benefits of aggressive compression while approaching the
deployability of hardware-native \FPeight{} KV caching.

%% file: sections/06c_limitations.tex
\section{Limitations}
\label{sec:limitations}
We present the three following limitations:
\begin{itemize}
  \item[1.] Although we deploy a single constant $c = 0.156$ for all models and heads, there may be 
  a more efficient constant that can be calibrated layer-wise for better accuracy. We also do not 
  consider the case that the constant is chosen depending on the distribuition of activations
  We ommit this out of simplicty and leave this for future work.
   \item[2.] UltraQuant's benefits in speed versus \FPeight{} are only observed when the context length is long enough, 
   or concurrency high enough, to exceed the resident cache capacity of the \FPeight{} baseline; for shorter context 
   lengths and lower concurrency, the benefits of the algorithm are not realized. Note that this is not the case when 
   compared to Ultra-TQ, where performance benefits are always realized. Larger clusters, other serving frameworks, 
   and non-CDNA4-optimized performance are also outside the scope of this evaluation.
   \item[3.] Ultra-TQ style centroid calibration is only applied to 10\% of layers for a small benefit; on most layers, such centroid
   calibration does not seem to provide any benefit. 
\end{itemize}

%% file: sections/05_kernel_implementation.tex
\section{Kernel Implementation in vLLM}
\label{sec:kernels}

We implement two serving endpoints in vLLM: \emph{Ultra-TQ}, an
optimized \TQ{} kernel that preserves the codebook representation, and
\emph{UltraQuant}, an FP4 approximation path in which the KV cache is
stored as raw FP4 micro-tensors with UE8M0 group scales and the
dequantization is performed by the matrix core itself.

The implementation targets the CDNA4 scaled-MFMA family, especially
\texttt{v\_mfma\_scale\_f32\_16x16x128\_f8f6f4}. The QK product reads
FP4 keys, FP8 queries, and UE8M0 scales together in one instruction.
The V path similarly avoids a software LUT by converting FP4 codes and
UE8M0 scales through hardware-supported low-precision operations.
Relative to a \TQ{} codebook kernel, the key systems change is removing
dequantization from the critical path: the query operand is pre-rounded
to FP8 outside the K-tile inner loop, and the grouped scale layout used
by the cache is arranged to match the MFMA scale operand layout.
\Cref{alg:fp4-constopt} gives the UltraQuant cache-encoding path; the
codebook-based Ultra-TQ encoding it replaces is in
\cref{app:lut-int4}.

\begin{algorithm}[t]
  \caption{FP4-ConstOpt KV encoding (UltraQuant)}
  \label{alg:fp4-constopt}
  \begin{algorithmic}[1]
    \Require K, V (BF16); rotation $R$; group size $g{=}32$; E2M1
      levels $F$; global constant $c$ (\cref{sec:uq-constopt}).
    \Statex \textbf{Write:}
    \State $K \gets R \cdot K$ \Comment{rotate keys; V unrotated}
    \State split $K, V$ into 32-channel groups
    \State \textbf{per group:} $m \gets \max_i |x_i|$;\;
      $E \gets \mathrm{round}(\log_2(c \cdot m))$
      \Comment{UE8M0 exponent; no per-token norm, no search}
    \State \textbf{per coord:} $\text{code} \gets q(x_i / 2^{E})$
      \Comment{$q$: round to nearest level in $F$}
    \State store 2 FP4 codes/byte $+$ 1 UE8M0 byte ($E$) per group
    \Statex \textbf{Decode:}
    \State $Q \gets R \cdot Q$;\; $Q \gets \mathrm{cast\_fp8\_e4m3}(Q)$
    \State load codes $+$ UE8M0 byte;\;
      $\hat{x} \gets \text{code} \cdot 2^{(\text{byte}-127)}$
    \State QK via scaled FP8$\times$FP4 MFMA; PV via FP4/scale path
    \Statex \textbf{Output:} hardware-native FP4+UE8M0 KV cache; dequant
      folds into the MFMA scale operand.
  \end{algorithmic}
\end{algorithm}

\subsection{Ultra-TQ Kernel Ladder}
\label{sec:ultra-tq}

\TQ{} compresses the KV cache to 4 bits per element, storing keys and
values in a codebook format. The cache is dequantized on the fly,
making decode memory-bandwidth-bound. Our kernels are inspired by
FLUTE~\citep{guo2024flute}, which restructures quantized matrices
offline to reduce bit-manipulation overhead during unpacking and uses
vectorized LUT access to reduce shared-memory pressure. We adapt these
ideas to the vLLM decode-attention path by adding GQA-aware tiling, a
structure-of-arrays KV cache layout, and AMD-specific MFMA dispatch.

The Ultra-TQ optimization ladder has three tiers. First, an improved
Triton kernel fixes the GQA, LUT, and layout pathologies of the
open-source \TQ{} baseline. Second, a native-ISA implementation issues
matrix-core instructions through AMD GCN intrinsics, giving direct
control over operand layout and instruction selection. Third, a
FlyDSL~\citep{li2026flydsl} implementation JIT-compiles to AMD GCN
ISA without per-device source files and exposes MFMA operand layouts
so we can select wider MFMA variants.

\begin{figure}[t]
  \centering
  \includegraphics[width=\columnwidth]{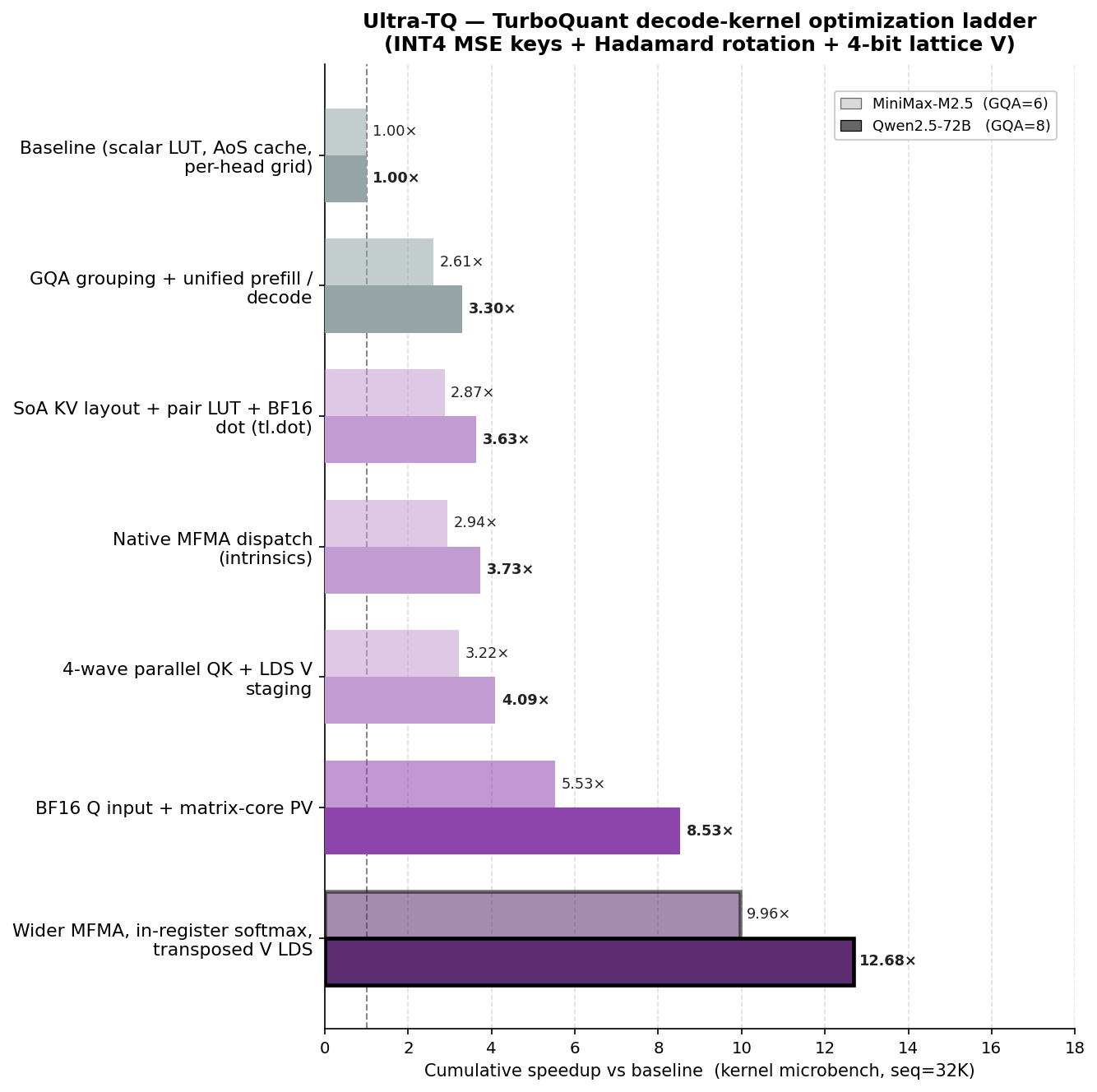}
  \caption{Ultra-TQ optimization ladder for the decode-attention
    kernel.}
  \label{fig:ultra-tq-ladder}
\end{figure}

\subsection{UltraQuant Kernel Ladder}
\label{sec:flydsl}

The UltraQuant kernel takes the FP4 plus UE8M0 cache layout described
in \cref{sec:ultra-quant} and adds three CDNA4-native
operations to the same DSL implementation. First, the QK product is
issued through the scaled F8F6F4 MFMA
(\texttt{v\_mfma\_scale\_f32\_16x16x128\_f8f6f4}), which reads FP4
keys, FP8 queries, and their UE8M0 scales together in one instruction;
BF16 keys never have to be materialized. Second, the V-dequant LUT is
replaced by a hardware conversion path that turns FP4 codes and UE8M0
scales directly into BF16. This removes software dequantization from
the critical path. Third, the FP8 query operand is lifted out of the
K-tile inner loop so every tile reuses the same pre-quantized query,
reducing redundant packing and easing register pressure.

\begin{figure}[t]
  \centering
  \includegraphics[width=\columnwidth]{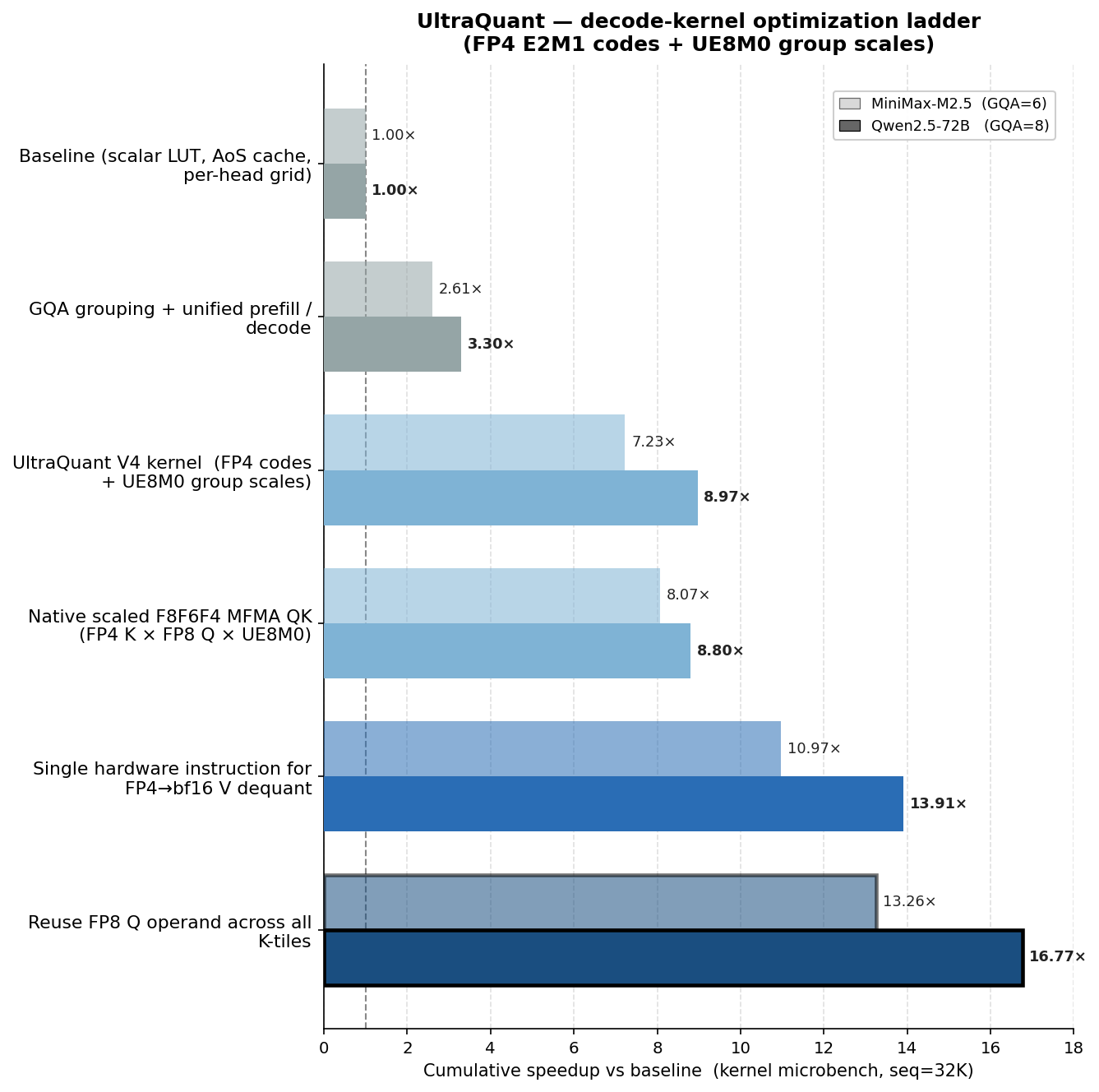}
  \caption{UltraQuant optimization ladder for the FP4 decode-attention
    kernel.}
  \label{fig:ultraquant-ladder}
\end{figure}

The implementation runs in the standard ROCm/vLLM stack. Ultra-TQ's
FlyDSL decode kernel has been upstreamed into vLLM mainline. We plan
to release the UltraQuant kernel as a follow-up contribution.

\subsection{Decode-Attention Kernel Counters}
\label{app:roofline-detail}

This supplements the roofline analysis in \cref{sec:roofline}.
\FPeight{} KV is bandwidth-bound and closest to the roof, but requires
$2\times$ the KV-cache space, halving the available capacity before
eviction. Ultra-TQ reaches 4-bit intensity but is dequant-bound: its
limiter is software codebook decode, an LDS gather, and a VALU FMA per
element, so in the worst case the compute cost outweighs the bandwidth
saving. \UQ{} reaches 4-bit intensity and stays closer to
bandwidth-bound, ${\sim}2.5\times$ higher than Ultra-TQ on the
roofline, because FP4 E2M1 with UE8M0 scales is consumed natively by
the scaled MFMA with no codebook, gather, or per-element FMA.

\Cref{tab:kernel-comparison} reports the corresponding decode-attention
kernel counters at $32$K context, batch $8$. The counters confirm that
4-bit KV roughly halves HBM traffic; Ultra-TQ's software LUT decode is
compute-bound despite the smaller payload, while \UQ{} folds
dequantization into the CDNA4 scaled-MFMA path, retaining most of the
memory saving at near-\FPeight{} kernel time.

\begin{table}[t]
  \centering
  \scriptsize
  \resizebox{\columnwidth}{!}{%
  \begin{tabular}{lccc}
    \toprule
    \textbf{KV format} & \textbf{Width} & \textbf{HBM bytes/step} &
    \textbf{Decode attn.\ (32K, batch 8)} \\
    \midrule
    \FPeight{} KV & 8-bit & ${\sim}23$\,MB & $1071$\,ms ($1.00\times$) \\
    Ultra-TQ      & 4-bit & ${\sim}12$\,MB & $2624$\,ms ($2.45\times$) \\
    \UQ{}         & 4-bit & ${\sim}12$\,MB & $1268$\,ms ($1.18\times$) \\
    \bottomrule
  \end{tabular}
  }
  \caption{Decode-attention kernel comparison at $32$K context, batch
    $8$, MiniMax-M2.5 on MI355X. \UQ{} nearly halves HBM traffic
    relative to \FPeight{} KV while staying within $1.18\times$ its
    kernel time; Ultra-TQ's software dequant erases the bandwidth
    saving.}
  \label{tab:kernel-comparison}
\end{table}

%% file: appendix/D_per_block_ablation.tex
\section{Ablations}
\label{app:ablations}

\subsection{Per-Block Scale}
\label{app:per-block-ablation}

This ablation isolates the contribution of the per-block-absmax scale
from that of the calibrated codebook, on GPT-OSS-20B GPQA (medium
effort, single seed; $1\sigma \approx 1.7$\,pp). Two effects stand out.
First, switching the adaptation statistic from a single per-token
$\ell_2$ norm to a per-block absmax (groups of 32) is the load-bearing
change: it is what recovers accuracy, independent of the codebook.
Second, with the per-block scale in place the codebook itself barely
matters --- replacing the calibrated Lloyd magnitudes with a uniform
grid stays within single-seed noise. Together, these justify the
deployed design: keep the per-block scale, and use the fixed FP4 grid
in place of a calibrated, model-specific codebook.

\begin{table}[t]
  \centering
  \scriptsize
  \resizebox{\columnwidth}{!}{%
  \begin{tabular}{llcc}
    \toprule
    \textbf{Configuration} & \textbf{Adaptation} & \textbf{Codebook} &
    \textbf{GPQA} \\
    \midrule
    TQ-t4nc (production) & per-token $\ell_2$ & Lloyd (calib.) & $0.6503$ \\
    K+V Lloyd, per-token & per-token $\ell_2$ & Lloyd (calib.) & $0.6237$ \\
    LMPb full            & per-block absmax   & Lloyd (calib.) & $0.6559$ \\
    Variant E            & per-block absmax   & uniform (RTN)  & $0.6528$ \\
    \bottomrule
  \end{tabular}
  }
  \caption{Per-block scale ablation (GPT-OSS-20B GPQA, medium, single
    seed). Holding rotation on, the per-token~$\rightarrow$~per-block
    adaptation moves $0.6237 \rightarrow 0.6559$ on the both-Lloyd
    configs; with per-block scaling fixed, Lloyd versus uniform
    codebook ($0.6559$ vs.\ $0.6528$) is within noise. Codebook and
    adaptation are co-fit, so the swap is not a perfectly clean A/B,
    but the direction is consistent across variants.}
  \label{tab:per-block-ablation}
\end{table}

\subsection{Global Constant}
\label{app:global-constant-ablation}

\Cref{fig:codebook-distortion} quantifies the codebook approximation:
plain per-32-group FP4 sits above the analytical Lloyd--Max floor, but
the constOpt per-block scale closes most of that MSE gap while keeping
the hardware-native FP4 grid, across $d \in \{64, 128, 256\}$.

We also sweep the global scaling constant $c$ against the
\FPeight{} (8-bit) baseline, on the same evaluation. The default
$c = 0.156$ is the MSE-optimal value derived in
\cref{sec:uq-constopt}, and the data confirm it: it beats \FPeight{} by
$+4.4$\,pp. Raising $c$ to $0.195$ (the power-of-two-optimal value, see
\cref{sec:limitations}) gives up the gain entirely and lands exactly at
the \FPeight{} baseline, and $c = 1.0$ (raw absmax with no MSE
shrinkage) falls $4.3$\,pp below \FPeight{}. Accuracy is monotonic in
the distance from the MSE-optimal constant.

\begin{table}[t]
  \centering
  \scriptsize
  \resizebox{\columnwidth}{!}{%
  \begin{tabular}{lcc}
    \toprule
    \textbf{Scheme} & \textbf{Accuracy} & \textbf{vs.\ \FPeight{} baseline} \\
    \midrule
    \FPeight{} (8-bit baseline)     & $63.0\%$ & --- \\
    fp4 $c=0.156$ (default)         & $67.4\%$ & $+4.4$\,pp \\
    fp4 $c=0.195$                   & $63.0\%$ & $0.0$\,pp \\
    fp4 $c=1.0$ (no shrinkage)      & $58.7\%$ & $-4.3$\,pp \\
    \bottomrule
  \end{tabular}
  }
  \caption{Global-constant ablation. The default $c=0.156$ is
    MSE-optimal and the only setting that improves on \FPeight{};
    larger $c$ erases the gain and $c=1.0$ (raw absmax) regresses below
    the 8-bit baseline.}
  \label{tab:global-constant-ablation}
\end{table}

\subsection{Cache-Pressure (GMU) Regime}
\label{app:gmu-ablation}

\begin{figure}[t]
  \centering
  \begin{subfigure}{0.49\columnwidth}
    \centering
    \includegraphics[width=\columnwidth]{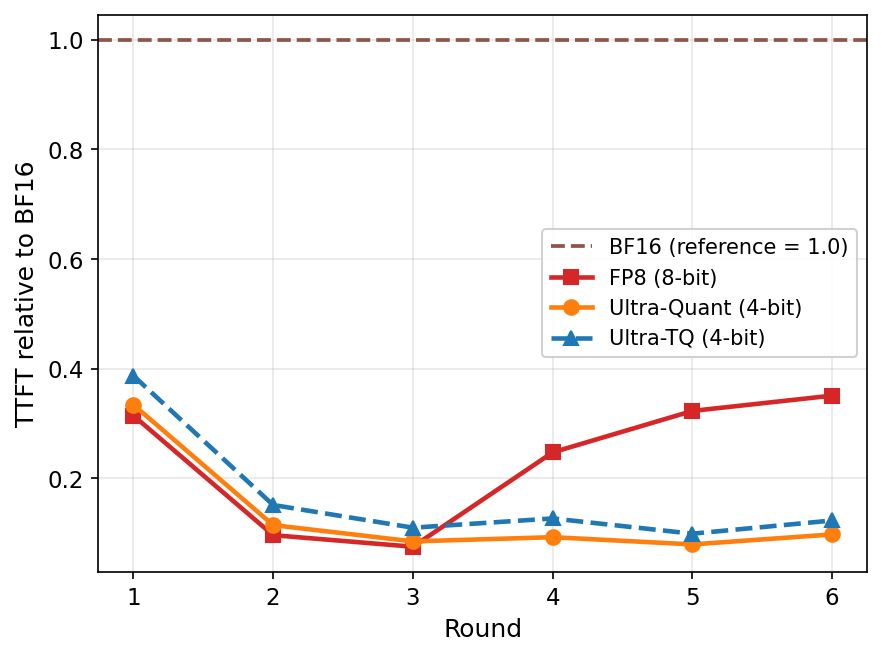}
    \caption{TTFT, GMU${=}0.60$.}
    \label{fig:gmu-ttft-60}
  \end{subfigure}
  \hfill
  \begin{subfigure}{0.49\columnwidth}
    \centering
    \includegraphics[width=\columnwidth]{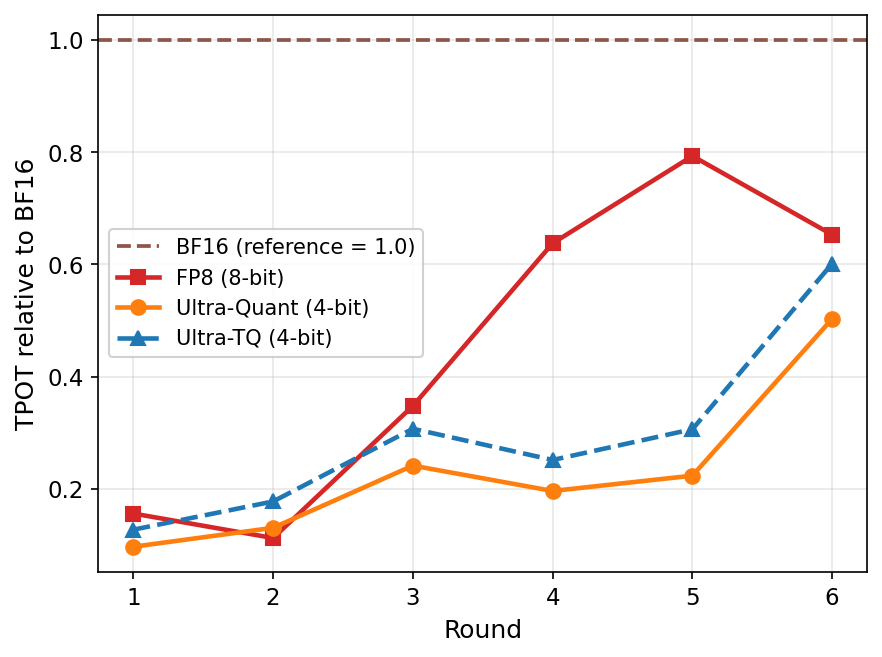}
    \caption{TPOT, GMU${=}0.60$.}
    \label{fig:gmu-tpot-60}
  \end{subfigure}
  \\[0.6em]
  \begin{subfigure}{0.49\columnwidth}
    \centering
    \includegraphics[width=\columnwidth]{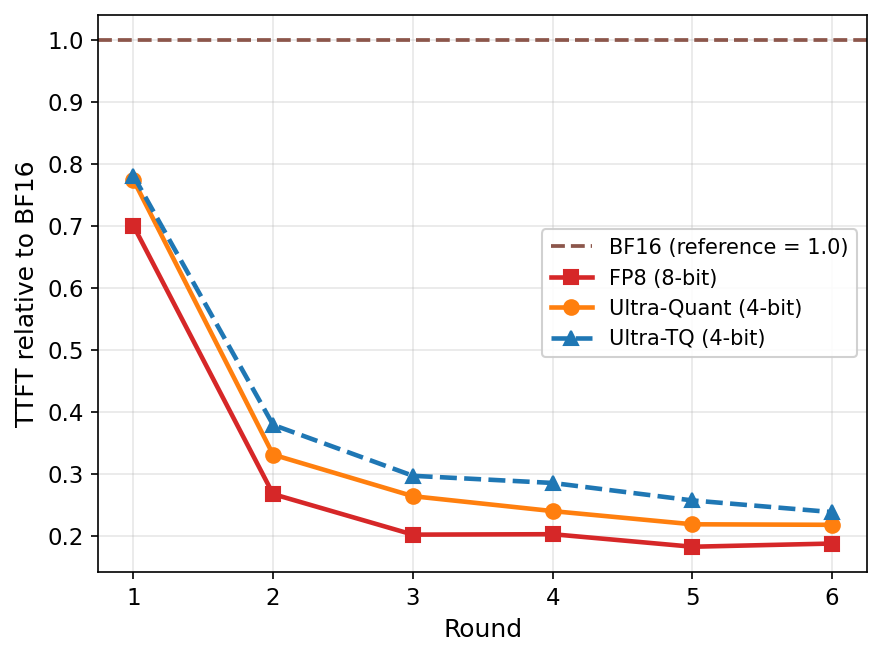}
    \caption{TTFT, GMU${=}0.65$.}
    \label{fig:gmu-ttft-65}
  \end{subfigure}
  \hfill
  \begin{subfigure}{0.49\columnwidth}
    \centering
    \includegraphics[width=\columnwidth]{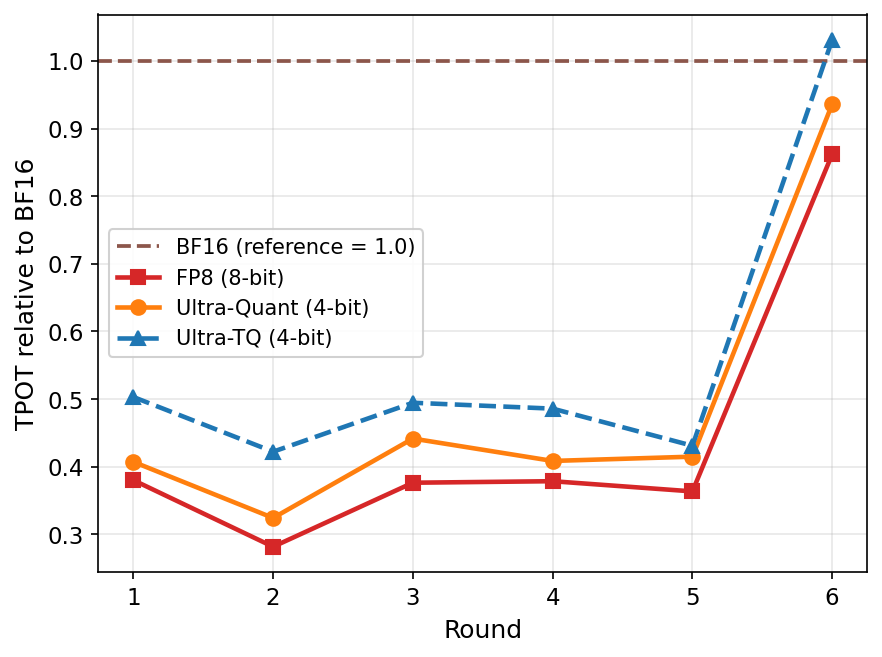}
    \caption{TPOT, GMU${=}0.65$.}
    \label{fig:gmu-tpot-65}
  \end{subfigure}
  \caption{Per-round serving latency (relative to \BFsixteen{}; lower
    is better) at two cache-pressure regimes. Top row: GMU${=}0.60$
    (the \cref{app:synthetic-multiturn} operating point), where
    \FPeight{} degrades in later
    rounds. Bottom row: GMU${=}0.65$, where the larger resident-cache
    budget keeps all three schemes close.}
  \label{fig:gmu-latency}
\end{figure}

We report the per-round latency at two GPU-memory-utilization (GMU)
operating points to show how the relative ranking shifts with
cache pressure. \Cref{fig:gmu-latency} stacks both regimes: the top row
is GMU${=}0.60$ (the operating point used in
\cref{app:synthetic-multiturn}) and the bottom row is
the lower-pressure GMU${=}0.65$. At GMU${=}0.60$ the resident-cache
budget is tight, so \FPeight{} degrades in the later rounds while \UQ{}
stays low; at GMU${=}0.65$ the budget is large enough that \FPeight{}
no longer degrades and all three schemes track closely. The
multi-turn ablation in \cref{app:synthetic-multiturn} uses the more
cache-pressured GMU${=}0.60$ regime, where the \UQ{} residency
advantage is largest.

\subsection{Concurrency, Input-Length, and Output-Length Sensitivity}
\label{app:sensitivity-ablations}

This section gives the full panel-level detail for the sensitivity
sweeps summarized in \cref{sec:sensitivity-ablations}, on MiniMax-M2.5
(TP$=2$, $2\times$MI355X).

\begin{figure}[t]
  \centering
  \includegraphics[width=\columnwidth]{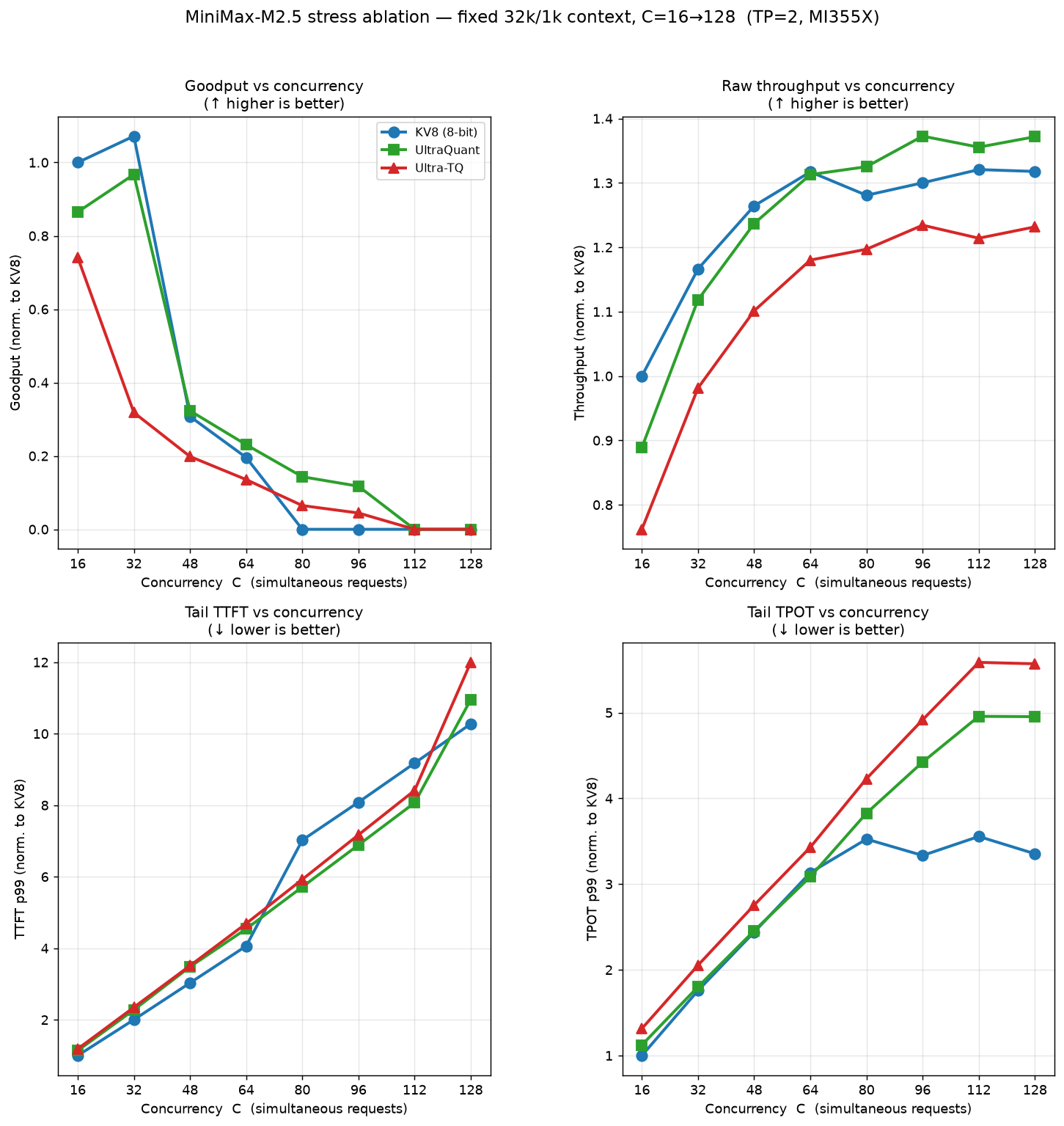}
  \caption{Full concurrency sweep on MiniMax-M2.5 (TP$=2$,
    $2\times$MI355X): goodput (top-left), raw throughput (top-right),
    P99 TTFT (bottom-left), and P99 TPOT (bottom-right) vs.\ load $C$
    for \FPeight{} KV, \UQ{}, and Ultra-TQ.}
  \label{fig:app-concurrency-full}
\end{figure}

\Cref{fig:app-concurrency-full} sweeps concurrency $C$. Goodput
(top-left) falls sharply for all schemes as load rises, with \UQ{}
sustaining it marginally longer than \FPeight{} KV. Raw throughput
(top-right) is close to tied between \FPeight{} KV and \UQ{}, with
Ultra-TQ trailing by ${\sim}10\%$ due to software dequant. P99 TTFT
(bottom-left) rises roughly linearly with $C$ and the three schemes
overlap. P99 TPOT (bottom-right) is the clearest separation:
\FPeight{} KV's tail TPOT plateaus because it caps admitted batch size
under queuing pressure, while the 4-bit schemes admit more concurrent
requests and pay a higher per-token cost, climbing to
${\sim}1.5$--$1.6\times$ that plateau.

\begin{figure*}[t]
  \centering
  \includegraphics[width=0.9\textwidth]{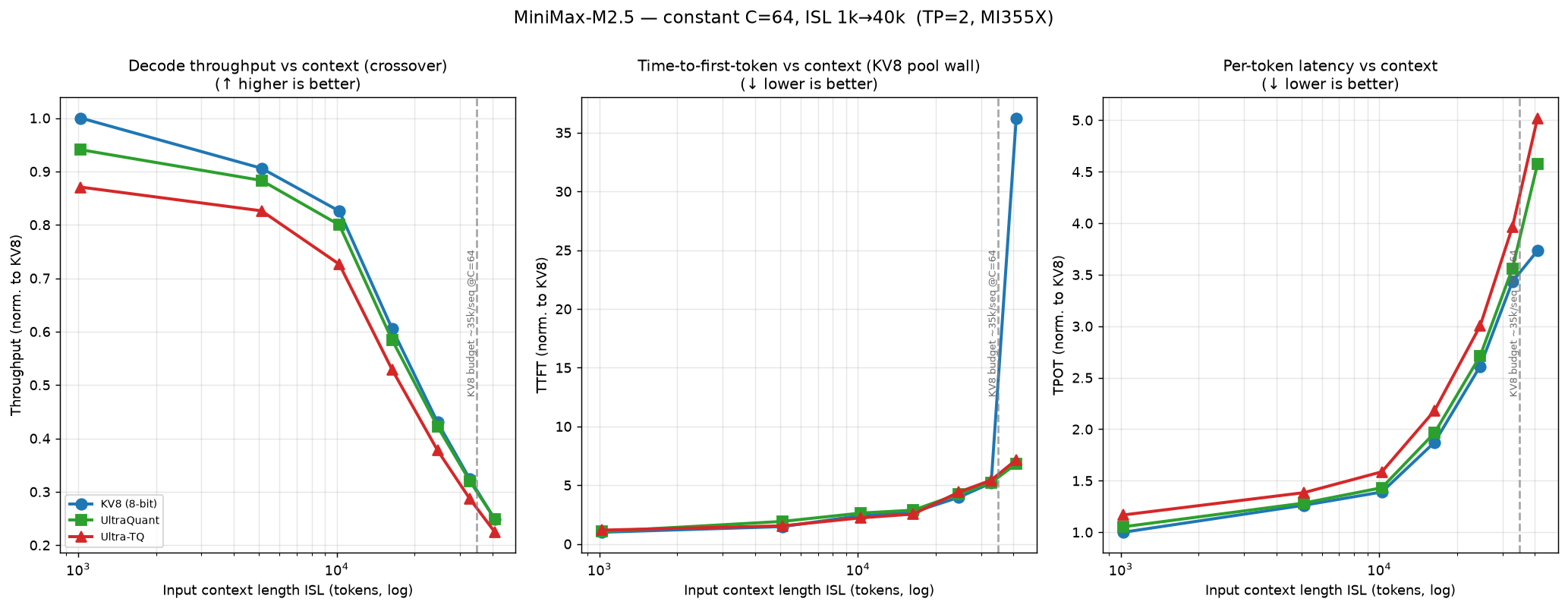}
  \caption{Full input-length sweep at fixed load on MiniMax-M2.5
    (TP$=2$, $2\times$MI355X): throughput and TTFT for \FPeight{} KV,
    \UQ{}, and Ultra-TQ across input context lengths.}
  \label{fig:app-input-size-full}
\end{figure*}

\Cref{fig:app-input-size-full} sweeps input context length at fixed
load. At short/medium context, \FPeight{} KV is fastest and the 4-bit
schemes pay a small tax (\UQ{} ${\sim}1$--$3\%$, Ultra-TQ
${\sim}7$--$10\%$). At $40$K context, \FPeight{} KV hits its resident
memory wall: it can no longer keep $64$ long sequences resident and
begins dropping/recomputing them. \UQ{}'s throughput pulls ahead past
this point, and \FPeight{}'s TTFT explodes to ${\sim}7\times$ its
short-context level while \UQ{} stays flat, ${\sim}5\times$ faster to
first token at the same setting. 4-bit KV buys capacity, not kernel
speed: it serves the long-context load that makes \FPeight{} KV fall
behind.

\begin{figure*}[t]
  \centering
  \includegraphics[width=0.9\textwidth]{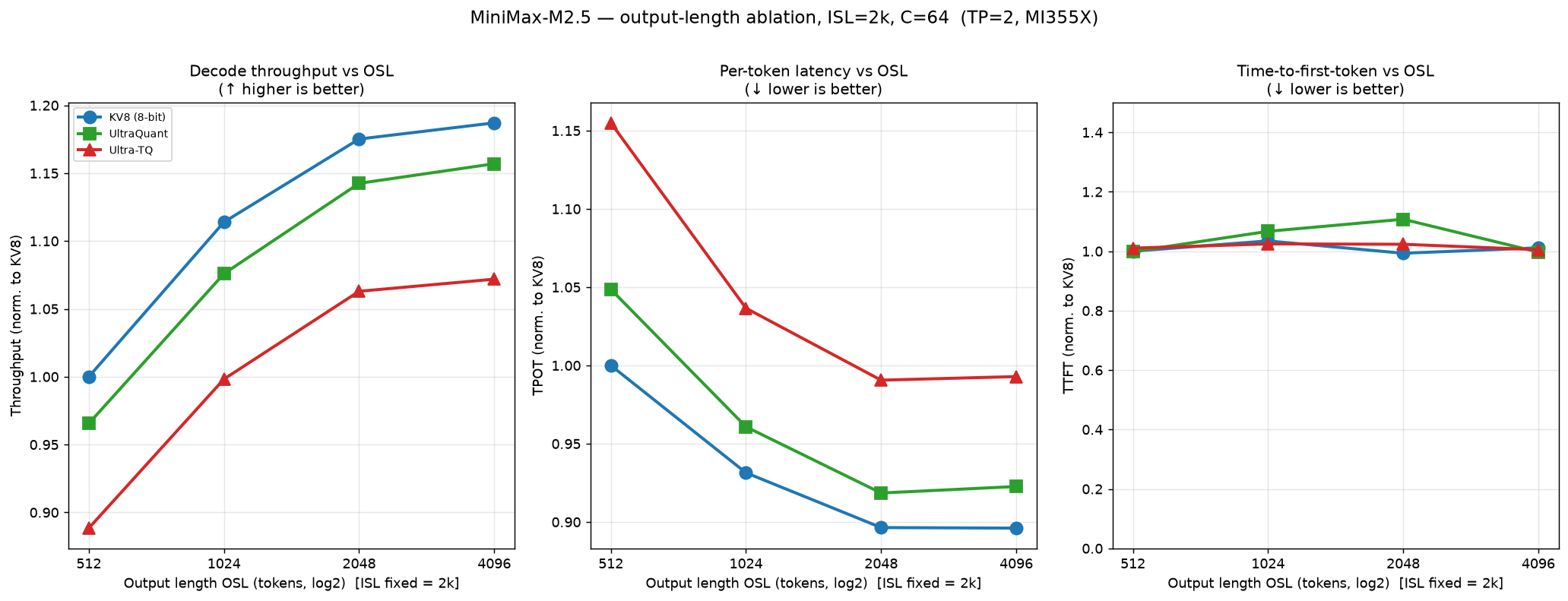}
  \caption{Full output-length sweep at fixed prompt size on
    MiniMax-M2.5 (TP$=2$, $2\times$MI355X): throughput (left), TPOT
    (middle), and TTFT (right) for \FPeight{} KV, \UQ{}, and
    Ultra-TQ.}
  \label{fig:app-output-size-full}
\end{figure*}

\Cref{fig:app-output-size-full} fixes the prompt and grows the answer
length, stressing pure decode. Throughput (left) rises and plateaus as
the fixed prefill cost is amortized over more output tokens; \UQ{}
shadows \FPeight{} KV to within $2$--$3\%$ at every output length.
TPOT (middle) starts higher and settles to a steady rate once decoding
is in full swing, with \UQ{} sitting a few percent above \FPeight{} KV
at a constant gap. TTFT (right) is flat and identical across all three
schemes, confirming output length affects decoding only.

\begin{figure*}[t]
  \centering
  \includegraphics[width=0.9\textwidth]{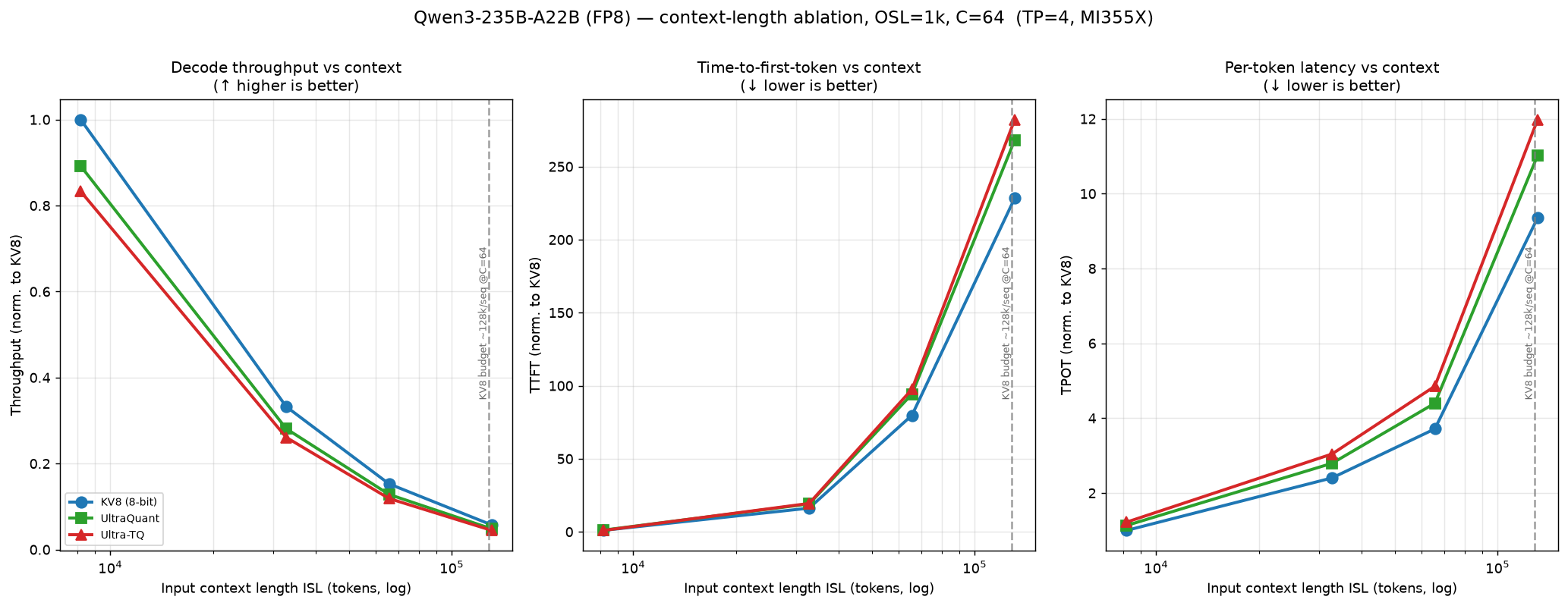}
  \caption{Input-length (context-length) sweep on Qwen3-235B-A22B
    (TP$=4$, fixed output length $1$K, $C{=}64$): decode throughput,
    TTFT, and per-token latency vs.\ context length for \FPeight{}
    KV, \UQ{}, and Ultra-TQ.}
  \label{fig:app-qwen235b-context}
\end{figure*}

We also ran the input-length ablation on Qwen3-235B-A22B
(\cref{fig:app-qwen235b-context}, TP$=4$, fixed output length
$1$K, $C{=}64$). It shows the same trend as MiniMax-M2.5 in
\cref{fig:app-input-size-full}: \UQ{} keeps up with \FPeight{} KV on
decode throughput, time-to-first-token, and per-token latency across
context lengths, while using half the KV memory.

%% file: appendix/G_accuracy_supplement.tex
\section{Additional Accuracy Results}
\label{app:accuracy-supplement}

This appendix supplements \cref{sec:accuracy-results} with the per-model
accuracy tables, per-task retrieval breakdowns, extended model validation,
and SWE-bench Lite cohort plots omitted from the main text for space.

\Cref{tab:accuracy-minimax,tab:accuracy-qwen,tab:accuracy-llama} report
\BFsixteen{} versus \UQ{} accuracy by model family, with benchmarks as rows,
for the retrieval and reasoning results discussed in
\cref{sec:accuracy-retrieval,sec:accuracy-reasoning}. Official RULER on
MiniMax~M2.5 is the largest single regression at $-6.95$; every other cell
sits within $2$~points of \BFsixteen{} in either direction.

\begin{table}[t]
  \centering
  \scriptsize
  \resizebox{\columnwidth}{!}{%
  \begin{tabular}{lrrr}
    \toprule
    \textbf{Benchmark} & \textbf{\BFsixteen{}} & \textbf{UltraQuant} &
    \textbf{$\Delta$~BF16} \\
    \midrule
    Official RULER   & \textbf{74.03} & 67.08 & $-6.95$ \\
    GPQA Diamond     & \textbf{83.84} & 81.82 & $-2.02$ \\
    GSM8K            & \textbf{88.86} & 87.11 & $-1.74$ \\
    AIME25 (5 seeds) & \textbf{79.17} & 78.33 & $-0.84$ \\
    \bottomrule
  \end{tabular}
  }
  \caption{MiniMax~M2.5 accuracy (\%).}
  \label{tab:accuracy-minimax}
\end{table}

\begin{table}[t]
  \centering
  \scriptsize
  \resizebox{\columnwidth}{!}{%
  \begin{tabular}{lrrr}
    \toprule
    \textbf{Benchmark} & \textbf{\BFsixteen{}} & \textbf{UltraQuant} &
    \textbf{$\Delta$~BF16} \\
    \midrule
    Official RULER$^{\dagger}$ & \textbf{99.38} & \textbf{99.38} & $0.00$ \\
    GPQA Diamond$^{\ddagger}$  & 51.01 & \textbf{56.57} & $+5.58$ \\
    GSM8K$^{\ddagger}$         & \textbf{95.53} & 95.22 & $-0.30$ \\
    AIME25$^{\ddagger}$        & \textbf{16.67} & \textbf{16.67} & $0.00$ \\
    \bottomrule
  \end{tabular}
  }
  \caption{Qwen accuracy (\%). $^{\dagger}$Qwen3.5-35B-A3B.
    $^{\ddagger}$Qwen3-32B.}
  \label{tab:accuracy-qwen}
\end{table}

\begin{table}[t]
  \centering
  \scriptsize
  \resizebox{\columnwidth}{!}{%
  \begin{tabular}{lrrr}
    \toprule
    \textbf{Benchmark} & \textbf{\BFsixteen{}} & \textbf{UltraQuant} &
    \textbf{$\Delta$~BF16} \\
    \midrule
    NIAH                   & 99.5 & \textbf{100.0} & $+0.50$ \\
    LongBench-E (13 tasks) & \textbf{54.19} & 53.97 & $-0.22$ \\
    GPQA Diamond           & \textbf{29.29} & 27.27 & $-2.02$ \\
    GSM8K                  & \textbf{78.09} & 77.03 & $-1.06$ \\
    \bottomrule
  \end{tabular}
  }
  \caption{Llama-3.1-8B-Instruct accuracy (\%).}
  \label{tab:accuracy-llama}
\end{table}

\begin{figure*}[t]
  \centering
  \includegraphics[width=\textwidth]{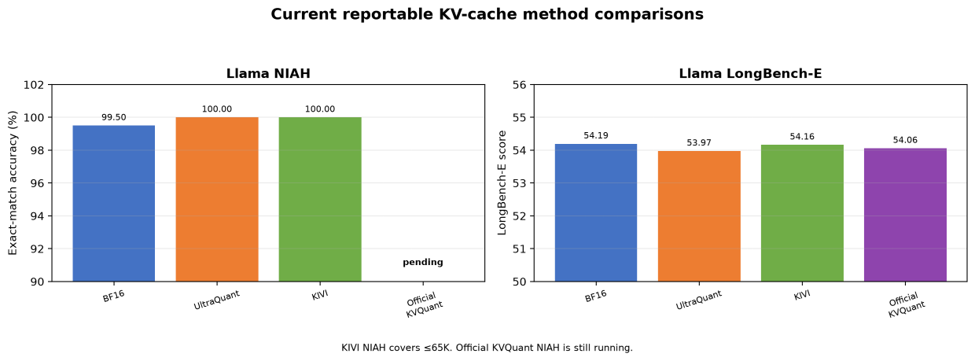}
  \caption{Llama-3.1-8B-Instruct NIAH and LongBench-E accuracy versus KIVI
    and official KVQuant. KIVI NIAH covers $\le 65$K context.}
  \label{fig:accuracy-kivi}
\end{figure*}

\Cref{fig:accuracy-kivi} compares KIVI and official KVQuant against
\BFsixteen{} and \UQ{} on Llama-3.1-8B-Instruct NIAH and LongBench-E.
KIVI tracks \BFsixteen{} through 65K context but OOMs at 131K; channel-wise
runtime quantization remains difficult because it quantizes across sequence
length during decode.

\begin{figure*}[t]
  \centering
  \includegraphics[width=\textwidth]{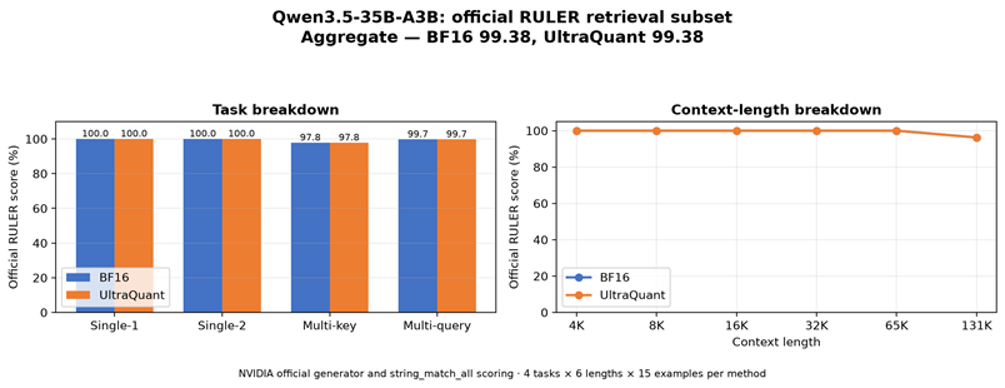}
  \caption{Official RULER retrieval subset for Qwen3.5-35B-A3B (four tasks
    $\times$ six lengths $\times$ 15 examples per method).}
  \label{fig:accuracy-ruler-qwen}
\end{figure*}

\Cref{fig:accuracy-ruler-qwen} breaks down the official RULER retrieval
subset for Qwen3.5-35B-A3B (four tasks $\times$ six lengths $\times$ 15
examples per method), where \UQ{} reaches exact parity with \BFsixteen{}
at $99.38$\% in aggregate (\cref{tab:accuracy-qwen}).

\begin{figure*}[t]
  \centering
  \includegraphics[width=\textwidth]{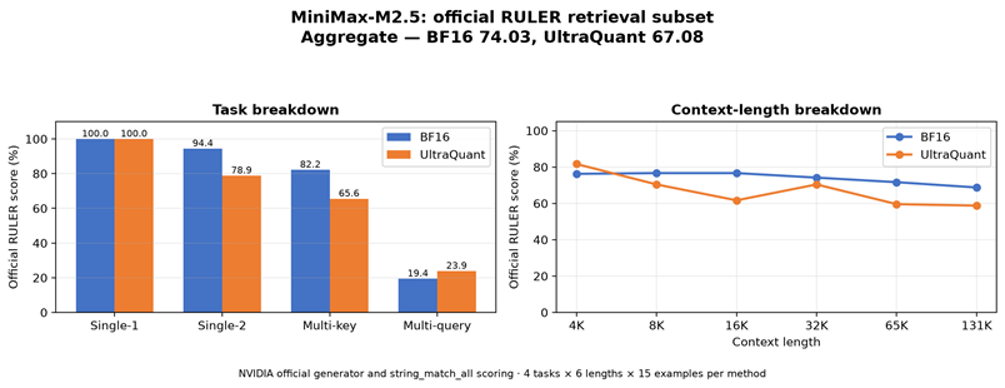}
  \caption{Official RULER retrieval subset for MiniMax~M2.5.}
  \label{fig:accuracy-ruler-minimax}
\end{figure*}

\Cref{fig:accuracy-ruler-minimax} gives the same RULER breakdown for
MiniMax~M2.5; loss concentrates in \texttt{single\_2} and
\texttt{multikey\_1}, with a gain on \texttt{multiquery}, matching the
$74.03$ vs.\ $67.08$ aggregate in \cref{tab:accuracy-minimax}.

\begin{figure*}[t]
  \centering
  \includegraphics[width=\textwidth]{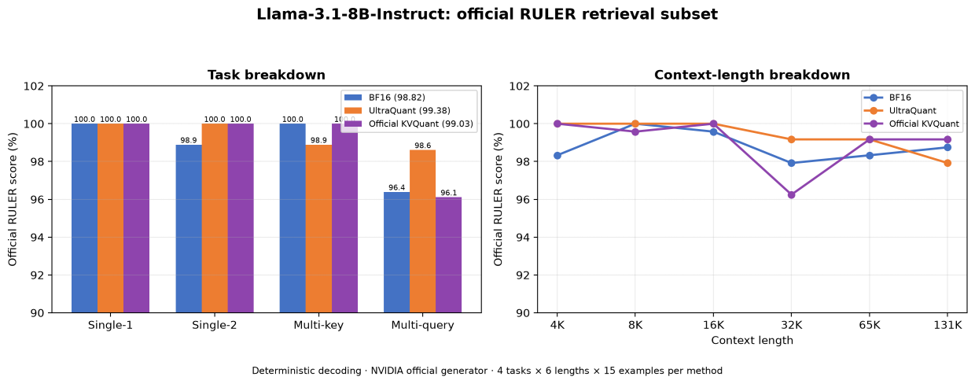}
  \caption{Official RULER retrieval subset for Llama-3.1-8B-Instruct,
    including official KVQuant.}
  \label{fig:accuracy-ruler-llama}
\end{figure*}

\Cref{fig:accuracy-ruler-llama} reports RULER for Llama-3.1-8B-Instruct,
including official KVQuant, contextualizing the NIAH and LongBench-E
summaries in \cref{tab:accuracy-llama}.

\begin{figure*}[t]
  \centering
  \includegraphics[width=\textwidth]{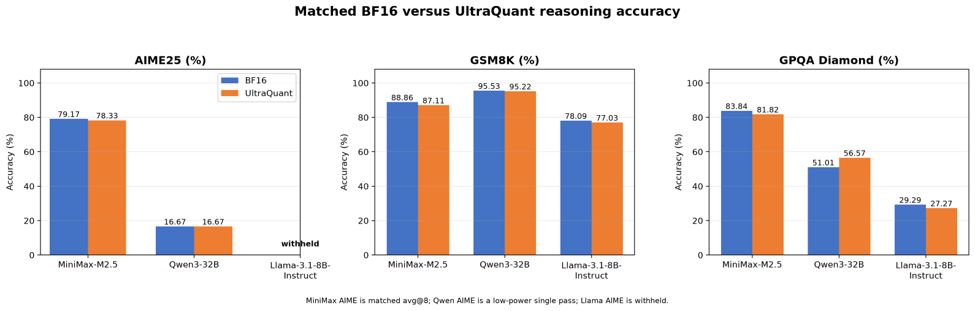}
  \caption{\BFsixteen{} versus \UQ{} accuracy on GPQA Diamond, GSM8K, and
    AIME25 (\cref{tab:accuracy-minimax,tab:accuracy-qwen,tab:accuracy-llama}).}
  \label{fig:accuracy-reasoning-bfuq}
\end{figure*}

\Cref{fig:accuracy-reasoning-bfuq} complements
\cref{tab:accuracy-minimax,tab:accuracy-qwen,tab:accuracy-llama} with
per-benchmark \BFsixteen{} versus \UQ{} panels on GPQA Diamond, GSM8K, and
AIME25; on MiniMax GPQA the intermediate schemes sit between the two reported
endpoints (\FPeight{} KV at $83.33$, Ultra-TQ matching \BFsixteen{} at
$83.84$).

\begin{figure*}[t]
  \centering
  \includegraphics[width=\textwidth]{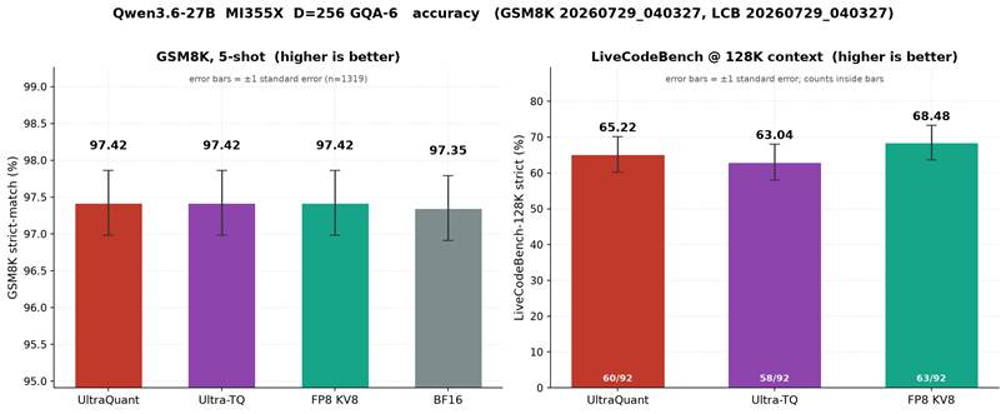}
  \caption{Qwen3.6-27B GSM8K (5-shot) and LiveCodeBench-128K accuracy on
    MI355X (TP$=2$). Error bars are $\pm 1$ standard error.}
  \label{fig:accuracy-qwen36}
\end{figure*}

\Cref{fig:accuracy-qwen36} validates Qwen3.6-27B on GSM8K (5-shot) and
LiveCodeBench-128K; error bars are $\pm 1$ standard error.

\begin{figure*}[t]
  \centering
  \includegraphics[width=\textwidth]{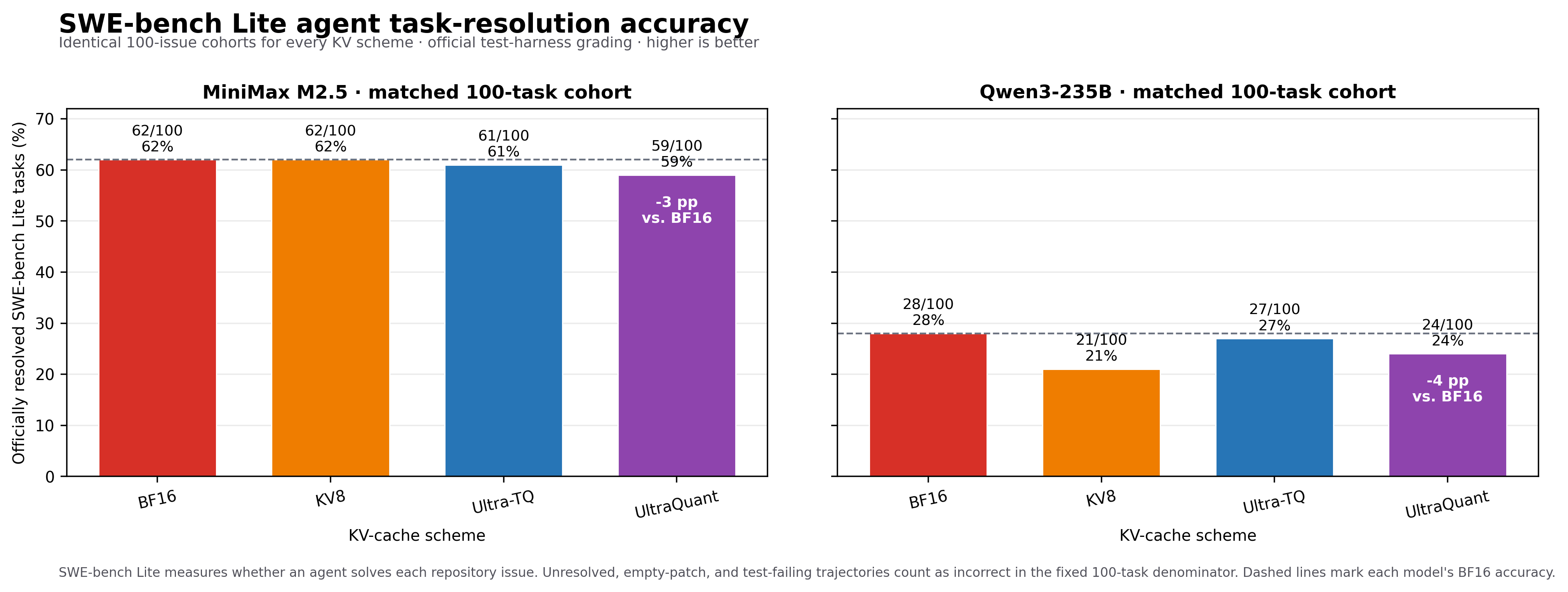}
  \caption{SWE-bench Lite agent task-resolution accuracy on fixed 100-task
    cohorts for MiniMax~M2.5 and Qwen3-235B (\cref{tab:accuracy-swebench}).}
  \label{fig:accuracy-swebench}
\end{figure*}

\Cref{fig:accuracy-swebench} plots resolved-task counts for MiniMax~M2.5 and
Qwen3-235B under all four KV schemes on the fixed 100-task SWE-bench Lite
cohort summarized in \cref{tab:accuracy-swebench}.

%% file: appendix/H_synthetic_multiturn.tex
\section{Multi-Turn Workload Ablation}
\label{app:synthetic-multiturn}

Beyond the production Claude Code replay in
\cref{sec:agentic-workloads}, we corroborate the same capacity effect
on a multi-turn workload that lets us set memory pressure directly. We
use the Gutenberg text corpus, from which vLLM synthesizes multi-turn
conversations in user/assistant role format under configurable prefix
and turn-length distributions. We run on MiniMax M2.5 and Qwen3-235B at
TP$=2$ on AMD MI355X, fixing concurrency at $C{=}32$ and reporting P50
TTFT and P50 TPOT per round.

\begin{figure}[t]
  \centering
  \begin{subfigure}{0.49\columnwidth}
    \centering
    \includegraphics[width=\linewidth]{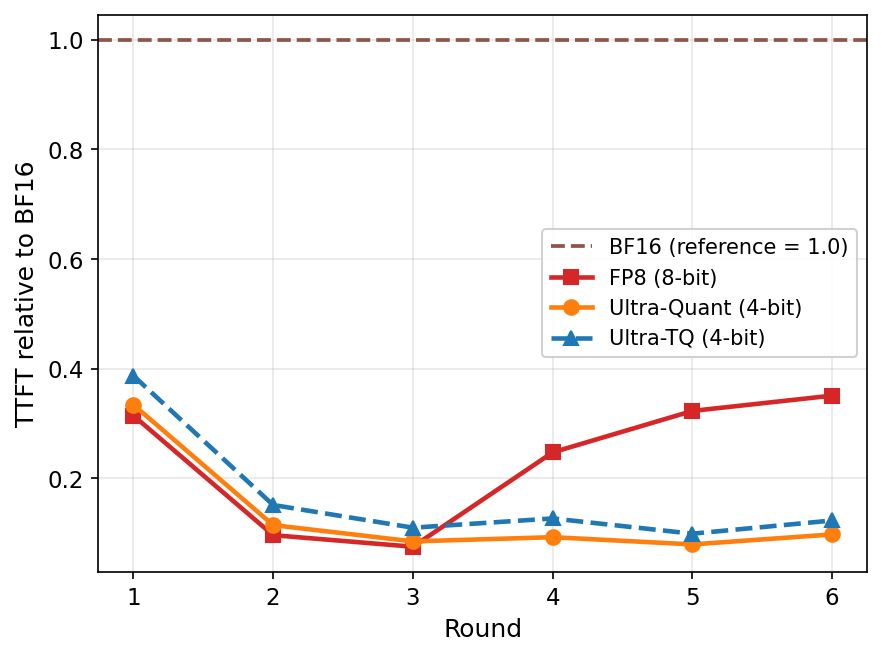}
    \caption{Time-to-first-token.}
    \label{fig:agentic-ttft}
  \end{subfigure}\hfill
  \begin{subfigure}{0.49\columnwidth}
    \centering
    \includegraphics[width=\linewidth]{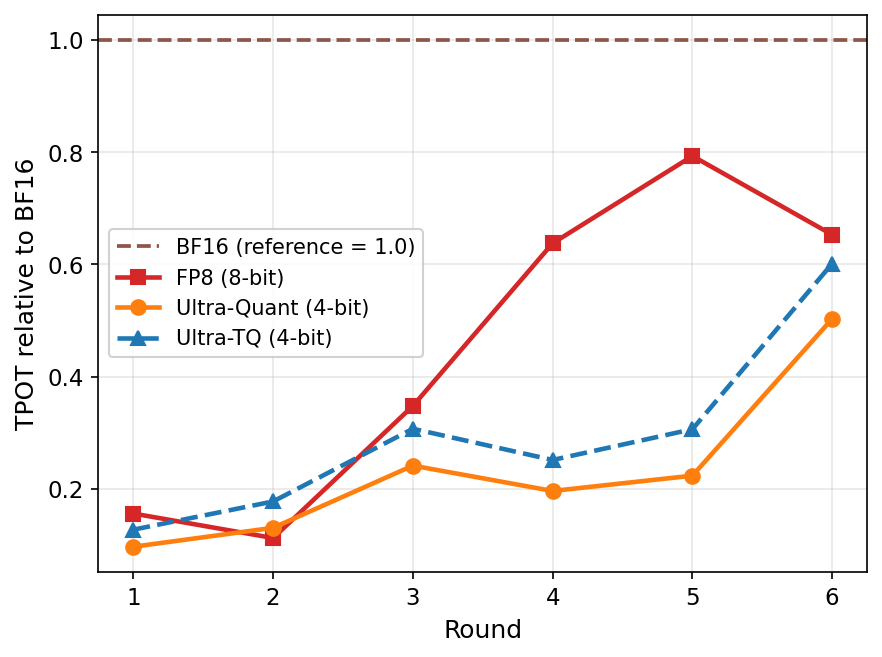}
    \caption{Time-per-output-token.}
    \label{fig:agentic-tpot}
  \end{subfigure}
  \caption{Per-round serving latency (relative to \BFsixteen{}; lower
    is better) on the Gutenberg multi-turn workload for \FPeight{},
    \UQ{}, and Ultra-TQ. \UQ{} keeps prefixes resident, so both
    (a)~time-to-first-token and (b)~time-per-output-token stay low
    across all rounds, while \FPeight{} degrades as accumulated context
    exceeds its resident-cache capacity.}
  \label{fig:agentic-latency}
\end{figure}

\Cref{tab:agentic-cache} and \cref{fig:agentic-latency} report this
workload on MiniMax M2.5. With long per-client prefixes and high
concurrency, \FPeight{} KV repeatedly evicts useful state while \UQ{}
keeps more prefixes resident. The advantage is largest in the late
rounds, where P50 TTFT improves $3.47\times$: \UQ{} holds both TTFT and
TPOT low across all six rounds, while \FPeight{} degrades as
accumulated context exceeds its resident-cache capacity.

\begin{table}[t]
  \centering
  \scriptsize
  \resizebox{\columnwidth}{!}{%
  \begin{tabular}{lc}
    \toprule
    \textbf{Metric} & \textbf{\UQ{} vs.\ \FPeight{} KV} \\
    \midrule
    P50 TTFT --- warm rounds (r2--3) & $0.86\times$ (\FPeight{} faster) \\
    P50 TTFT --- late rounds (r4--6) & $\mathbf{3.47\times}$ \\
    P50 TTFT --- all rounds          & $2.3\times$ \\
    Output throughput                & $\mathbf{1.63\times}$ \\
    \bottomrule
  \end{tabular}
  }
  \caption{Gutenberg multi-turn serving results on MiniMax-M2.5, TP$=2$,
    AMD MI355X, reported as \UQ{} relative to the \FPeight{} KV baseline
    (higher is better except where noted). The advantage appears in the
    late rounds, where long per-client prefixes exceed the effective
    resident-cache capacity of \FPeight{}: TTFT improves $3.47\times$
    and is recovered through cache residency rather than re-prefill.}
  \label{tab:agentic-cache}
\end{table}

%% file: appendix/F_claude_trace_adaptive.tex
\section{Claude Code Adaptive SLO Supplement}
\label{app:claude-trace-adaptive}

\begin{figure*}[t]
  \centering
  \begin{subfigure}{0.49\textwidth}
    \centering
    \includegraphics[width=\linewidth]{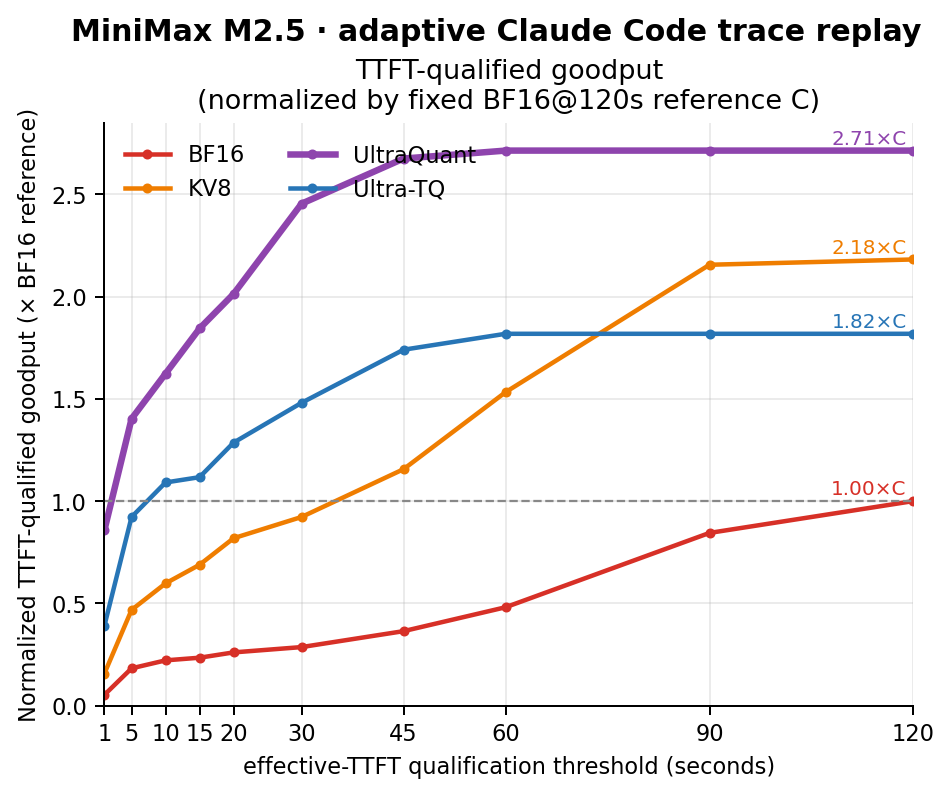}
    \caption{MiniMax M2.5.}
    \label{fig:app-trace-minimax-goodput}
  \end{subfigure}\hfill
  \begin{subfigure}{0.49\textwidth}
    \centering
    \includegraphics[width=\linewidth]{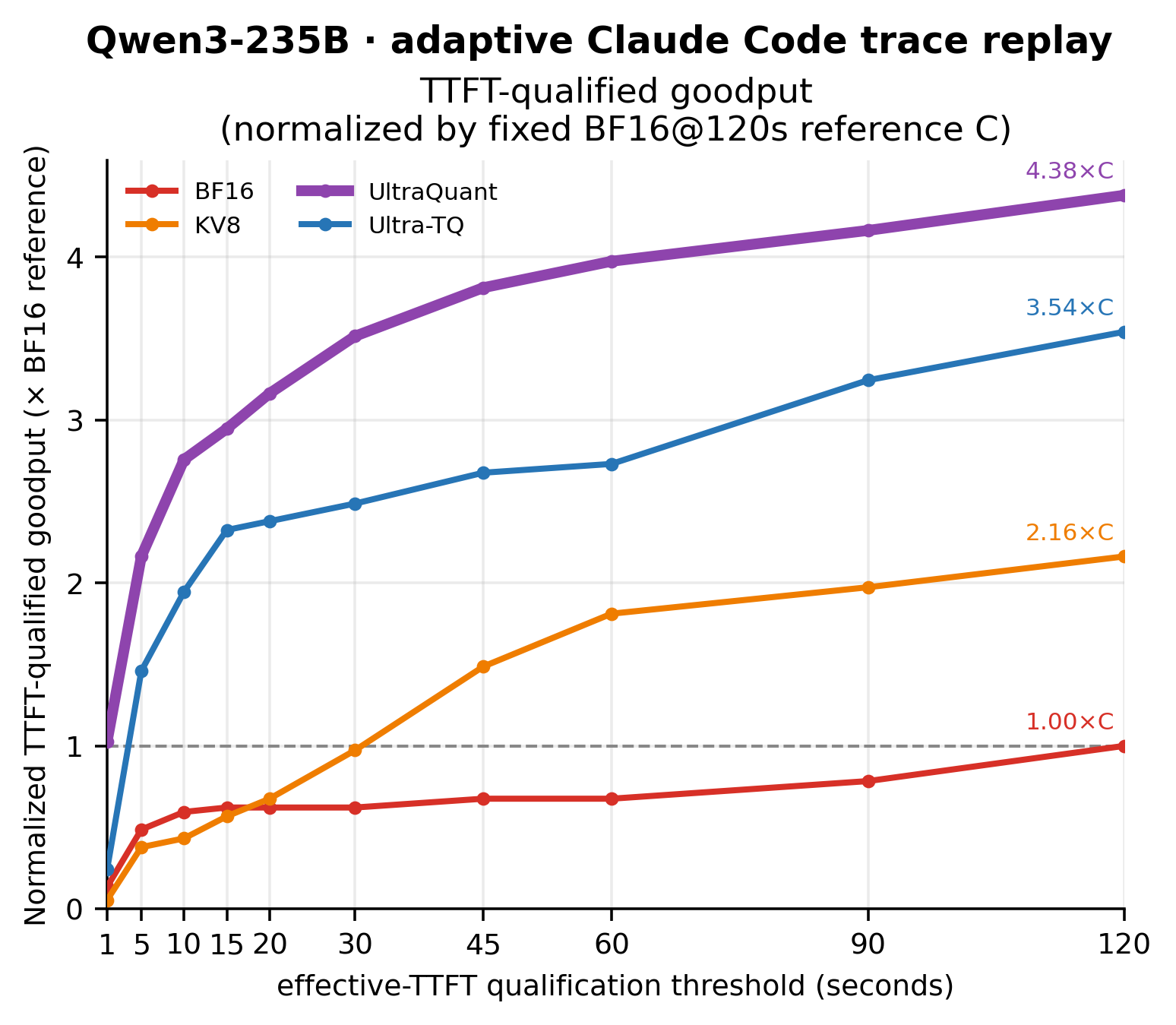}
    \caption{Qwen3-235B.}
    \label{fig:app-trace-qwen-goodput}
  \end{subfigure}
  \caption{TTFT-qualified completed-request goodput on the Claude Code trace
    replay (TP$=2$, AMD MI355X), normalized relative to \BFsixteen{}.}
  \label{fig:app-trace-goodput}
\end{figure*}

\begin{figure*}[t]
  \centering
  \begin{subfigure}{0.49\textwidth}
    \centering
    \includegraphics[width=\linewidth]{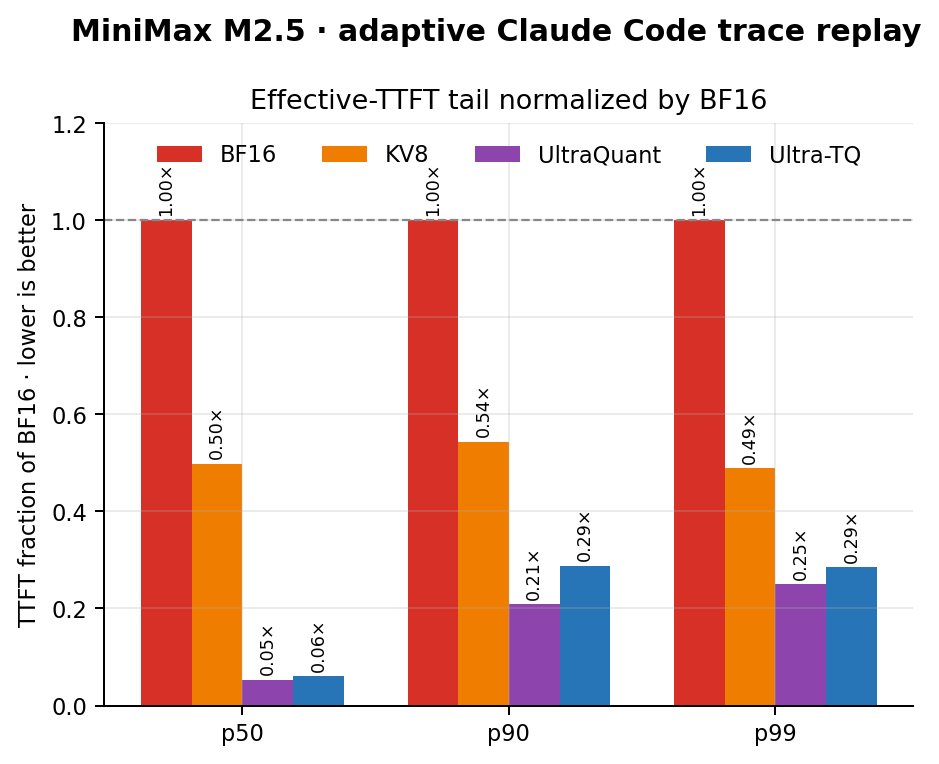}
    \caption{MiniMax M2.5.}
    \label{fig:app-trace-minimax-tail}
  \end{subfigure}\hfill
  \begin{subfigure}{0.49\textwidth}
    \centering
    \includegraphics[width=\linewidth]{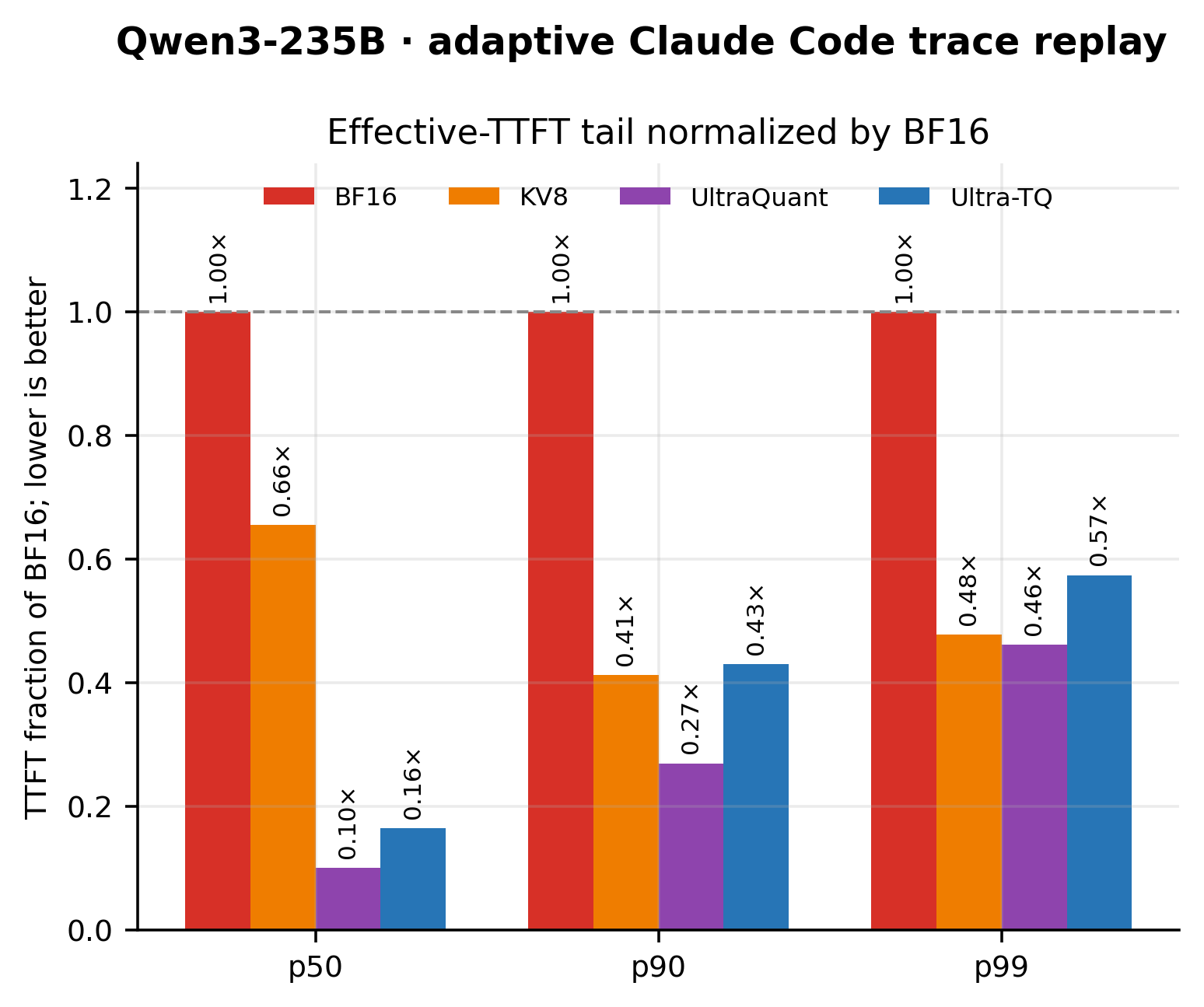}
    \caption{Qwen3-235B.}
    \label{fig:app-trace-qwen-tail}
  \end{subfigure}
  \caption{Effective-TTFT tail distributions (p50/p90/p95/p99) on the Claude
    Code trace replay (TP$=2$, AMD MI355X).}
  \label{fig:app-trace-effective-ttft}
\end{figure*}

This appendix supplements \cref{sec:agentic-workloads} with the remaining
adaptive-SLO panels from the production-derived Claude Code trace replay.
\Cref{fig:app-trace-goodput} reports TTFT-qualified completed-request goodput:
the rate of finished requests whose first token arrives within each swept
TTFT budget, normalized by the \BFsixteen{} qualified-request total.
\Cref{fig:app-trace-effective-ttft} reports effective-TTFT tail distributions
(p50/p90/p95/p99) at the same operating points. On MiniMax M2.5, \UQ{}
reaches $2.71\times$ \BFsixteen{} on both cumulative SLO-qualified served
requests (main text) and TTFT-qualified goodput; on Qwen3-235B, both metrics
reach $4.38\times$ \BFsixteen{}.

%% file: appendix/E_turboquant_algorithm.tex
\section{Calibrated LUT int4 KV Encoding (Ultra-TQ)}
\label{app:lut-int4}

For reference, \cref{alg:lut-int4} gives the codebook-based encoding
path that UltraQuant replaces. It is the TurboQuant-style scheme: a
calibrated 4-bit codebook per side with a per-token norm scalar on the
key side, dequantized through a lookup table on read. UltraQuant
(\cref{alg:fp4-constopt}) keeps the rotation but drops the codebook and
the per-token norm in favor of a fixed FP4 grid with a per-block UE8M0
scale, so the dequantization folds into the matrix core.

\begin{algorithm}[t]
  \caption{Calibrated LUT int4 KV encoding (Ultra-TQ)}
  \label{alg:lut-int4}
  \begin{algorithmic}[1]
    \Require K, V (BF16); rotation $R$; group size $g$; codebook $C_K$.
    \Statex \textbf{Write:}
    \State $K \gets R \cdot K$ \Comment{rotate keys; V unrotated}
    \State split $K, V$ into $g$-channel blocks
    \State K: store block norm; encode each coord to nearest $C_K$ index
    \State V: store block (scale, zero); encode each coord to 4-bit
      uniform
    \Statex \textbf{Decode:}
    \State $Q \gets R \cdot Q$
    \State load K/V codes $+$ metadata per tile
    \State K: dequant via $C_K \cdot \text{norm}$;\;
      V: dequant via $\text{scale} \cdot \text{code} + \text{zero}$
    \State compute QK and PV on dequantized $K, V$
    \Statex \textbf{Output:} rotated-codebook K $+$ uniform V cache;
      dequant paid on read.
  \end{algorithmic}
\end{algorithm}